\documentclass[letterpaper,journal,10pt]{IEEEtran} 
\usepackage{setspace} 

\usepackage{cite}

\usepackage{graphicx}
\DeclareGraphicsExtensions{.pdf,.jpg}

\usepackage[cmex10]{amsmath}
\usepackage{amsfonts}

\usepackage[noend]{algpseudocode}

\usepackage{array}
\usepackage{tikzpagenodes}

\usepackage{url}
\usepackage{fancyhdr}
\usepackage[left=1cm,right=1cm,top=1.5cm,bottom=1.8cm]{geometry}
\begin{document}

\title{Linguistic Descriptions for Automatic Generation of Textual Short-Term Weather Forecasts on Real Prediction Data}

\author{\IEEEauthorblockN{A. Ramos-Soto}, \IEEEauthorblockN{A. Bugar\'in}, \IEEEauthorblockN{S. Barro}, and \IEEEauthorblockN{J. Taboada}
\thanks{A. Ramos-Soto, A. Bugar\'in and S. Barro are with the Research Centre on Information Technologies (CiTIUS), University of Santiago de Compostela, Spain (e-mail: alejandro.ramos@usc.es, alberto.bugarin.diz@usc.es, senen.barro@usc.es). J. Taboada is with MeteoGalicia, Santiago de Compostela, Spain (e-mail: juan.taboada@meteogalicia.es).}
\thanks{This work was supported by the Spanish Ministry for Economy and Competitiveness under grant TIN2011-29827-C02-02. It was also supported in part by the European Regional Development Fund (ERDF/FEDER) under the project CN2012/151 of the Galician Ministry of Education.}
\thanks{A. Ramos-Soto is supported by the Spanish Ministry for Economy and Competitiveness (FPI Fellowship Program).}}

\maketitle
  \begin{tikzpicture}[remember picture,overlay]
    \node[align=center,text=black] at ([yshift=1cm]current page text area.north) {\small This article has been accepted for publication in a future issue of this journal, but has not been fully edited. Content may change prior to final publication.\\ \small Citation information: DOI 10.1109/TFUZZ.2014.2328011, IEEE Transactions on Fuzzy Systems};
    \node[align=center,text=black] at ([yshift=-1cm]current page text area.south) {\small 1063-6706 (c) 2013 IEEE. Personal use is permitted, but republication/redistribution requires IEEE permission. \\ \small See \url{http://www.ieee.org/publications_standards/publications/rights/index.html} for more information.};
  \end{tikzpicture}%

\begin{abstract}
We present in this paper an application which automatically generates textual short-term weather forecasts for every municipality in Galicia (NW Spain), using the real data provided by the Galician Meteorology Agency (MeteoGalicia). This solution combines in an innovative way computing with perceptions techniques and strategies for linguistic description of data together with a natural language generation (NLG) system. The application, named GALiWeather, extracts relevant information from weather forecast input data and encodes it into intermediate descriptions using linguistic variables and temporal references. These descriptions are later translated into natural language texts by the natural language generation system. The obtained forecast results have been thoroughly validated by an expert meteorologist from MeteoGalicia using a quality assessment methodology which covers two key dimensions of a text: the accuracy of its content and the correctness of its form. Following this validation GALiWeather will be released as a real service offering custom forecasts for a wide public.
\end{abstract}


\section{Introduction}
In recent years, governments and agencies from many countries have increasingly focused efforts on improving the accessibility of their citizens to public data, i.e., all the data that public bodies in a given country produce, collect or pay for, which is widely known as the Open Data paradigm \cite{bib_UE}. These resources, which come from many different fields of knowledge, offer a high potential for re-use in new products and services. This scenario has been described very graphically with the following statement ``data is the new oil for the digital age'' \cite{bib_oil}.

However, there is still a significant gap between the resources offered by public institutions and the necessities of their potential consumers. One reason is that the publishing bodies are usually focused on the availability of their datasets rather than on providing tools or means for accessing and processing them. This often results in extensive catalogues of heterogeneous data which have almost no direct value for the potential consumers of that data.

Besides a lack of standardization, there is also a lack of tools and services which allow a better access and comprehension of the raw data provided by the public institutions. An interesting and illustrative example of this kind of services can be found in meteorology, where meteorological agencies offer both raw data and also several types of information pieces (such as forecasts, reports or meteorological warnings) that are elaborated by meteorologists from these raw data.

Artificial Intelligence provides us with tools which allow us to process and understand this massive availability of huge quantities of data. Originally, this objective has been assumed by the knowledge discovery in databases (KDD) field, but more specifically by its core stage, the data mining field \cite{bib_kdd}, which assembles several tasks such as classification, association, clustering, trend analysis or summarization \cite{bib_datamining}. Summarization is of particular interest, since it abstracts data into useful information at different levels and dimensions. The abstracted information can adopt many forms, although the most common services come in the form of web-based visualization tools. However, other approaches taken by research fields such as natural language generation (NLG) or soft computing offer solutions to convert and summarize data into textual information which can be easily consumed by human users.

The creation of automatic textual summaries of data is a task which originally started within the NLG field. Several NLG approaches which generated summaries of data include ANA \cite{nlg_ANA}, which generated summaries of stock market activity; LFS \cite{nlg_LFS}, which generated summaries of statistical data; SUMGEN \cite{nlg_SUMGEN}, which generated summaries of events in a battle simulation; TEMSIS \cite{nlg_TEMSIS}, which generated summaries of environmental data; TREND \cite{nlg_TREND}, which generated summaries of historical weather data; and, more recently, BabyTalk \cite{bib_neonatal}, which generates medical reports for neonatal intensive care data. However, the most successful NLG systems for data summarization, at least in terms of public impact and usefulness, generate automatic textual weather forecasts from numerical prediction data.  A few systems, such as FoG \cite{nlg_goldberg}, MultiMeteo \cite{nlg_multimeteo} and SUMTIME-MOUSAM \cite{bib_mousam}, \cite{nlg_mousam}, have been used by meteorological agencies to automatically produce public weather forecasts.

At the same time, within the fuzzy logic and soft computing field, the paradigm of computing with words (CWW) \cite{bib_flcwwzadeh}, and its later evolution computing with perceptions (CWP) \cite{bib_tcpzadeh}, \cite{bib_mendel}, made their appearance in the 1990s. As opposed to other classical approaches, these paradigms involve a fusion of natural languages and computation with linguistic variables \cite{bib_tcpzadeh}. Although many new approaches based on CWW have emerged, one of the most promising tools is linguistic data summarization \cite{bib_yagerfc90}, \cite{bib_kacprzykproto}, which employs fuzzy quantified propositions to obtain linguistic summaries involving one variable (as in ``Most of the dogs are brown'' or ``A few trees are tall'') or more than one variable (as in ``Some of the brown dogs are heavy'' or ``Most of the tall trees are very old'').
Since then, linguistic summarization from CWW has been applied in several practical cases and, with the appearance of CWP, some authors have started to refer to linguistic summaries as linguistic descriptions of data (LDD) \cite{bib_gracianmichio}, which understand linguistic summaries as a tool to describe human perceptions. For reasons of clarity, we will use in this paper the term linguistic descriptions of data. Examples of fields of application of linguistic description approaches include descriptions of the patient inflow in health centers \cite{bib_granada}, domestic electric consumption reports \cite{bib_consumption}, human activity based on mobile phone accelerometers \cite{bib_accelerometer} or human gait quality \cite{bib_albertogait}. Other approaches use more complex expressions involving relationships among different attributes (in economic data \cite{bib_kobayashi}, in sales data \cite{bib_kacprzyk} or the analysis of investment fund quotations \cite{bib_investment}).

Most of these approaches are very strongly dependent on the field of application and the users' needs of information. A more general approach which is able to construct different kinds of linguistic descriptions regardless of the application domain is still an open challenge in this field. Nevertheless, steps in this direction have been taken by providing general criteria on how to structure quantified sentences in order to obtain more complex descriptions \cite{bib_kacprzykhigh} or on how to build and evaluate linguistic descriptions \cite{bib_felixisda}, \cite{bib_rita2012}, \cite{bib_menendez}.

Another open challenge is the relationship between linguistic descriptions in CWP and NLG. Until now, both have followed separate paths, although it remains clear that both can contribute to each other in a substantial way \cite{bib_kacprzyk}.

With both linguistic descriptions from CWP and textual summaries from NLG as inspiration, we present in this paper GALiWeather \cite{bib_galiweather}, an application which automatically generates short-term weather forecasts in the form of natural language texts for the Galician Meteorological Agency (MeteoGalicia) \cite{bib_meteogalicia}. This solution employs in an innovative way a LDD computational method combined with a NLG system in order to solve a real life information need, as opposed to other approaches which only present test use cases and do not address the whole problem of adapting their solutions to real final user needs and demands. For this, the use of fuzzy procedures through linguistic variables and quantifiers allows the application to model imprecise concepts included in the linguistic descriptions. Furthermore, the quality of these descriptions, which are generated as natural language texts by the NLG system, has been assessed by an expert meteorologist in two key dimensions, verifying that the textual forecasts are both correct and properly expressed.

The next section introduces the context in which this solution has been devised. In Section III a formal description of the forecast input data and the linguistic description computational method is provided, followed by an extensive overview of the NLG system. Section IV addresses the validation process and results obtained for our application. Section V contains some insights about a methodological conceptualization of our approach and finally in Section VI we present the most relevant conclusions.

\section{Short-term web forecasts for Galicia}
\begin{figure}[!t]
\centering
\includegraphics[width=0.8\columnwidth]{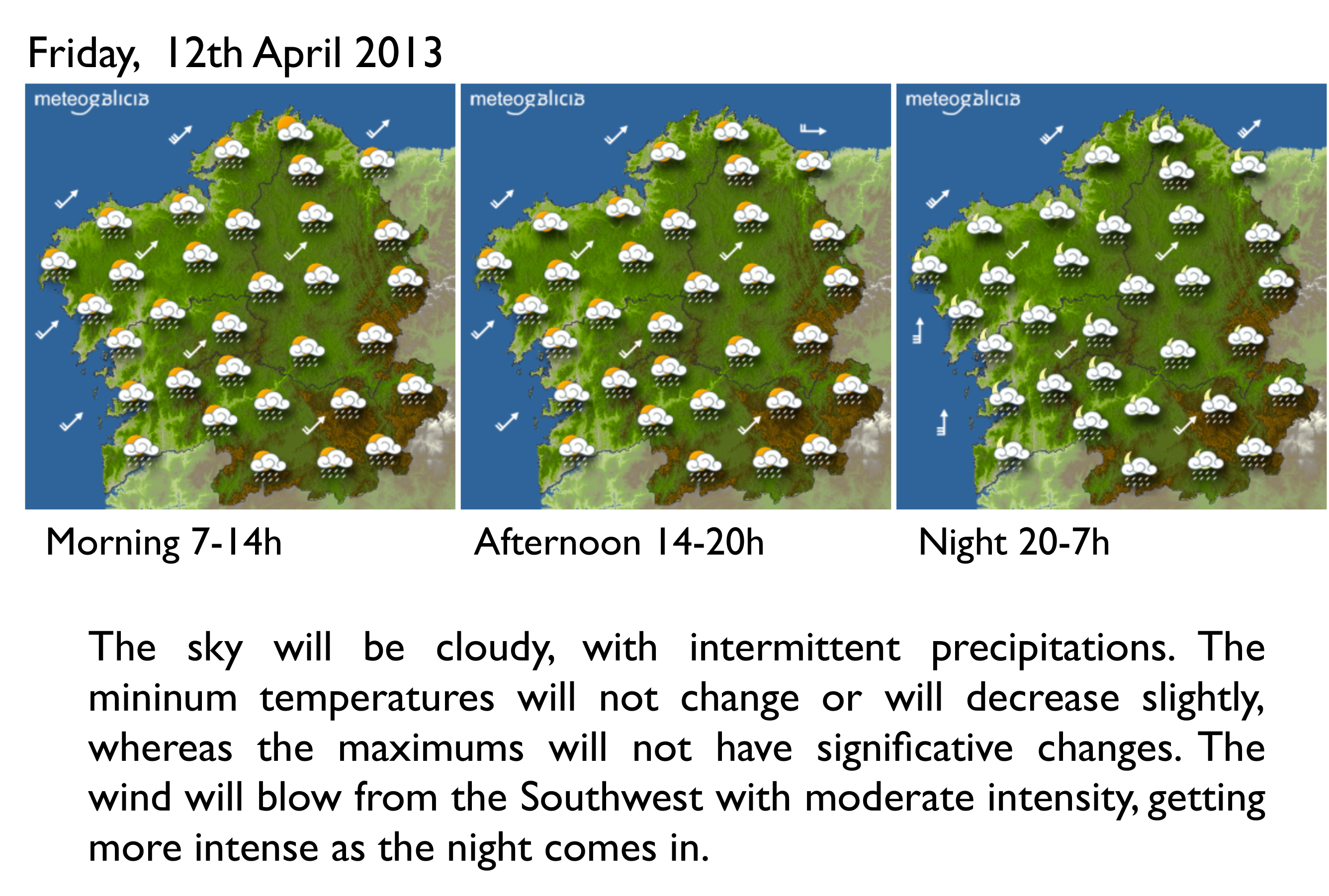}
\caption{Example of a real weather forecast for 12th April, 2013 for Galicia, published at \cite{bib_meteogalicia}.} 
\label{forecastapril}
\end{figure}

The operative weather forecasting offered by the Galician (NW Spain) Meteorology Agency through its website (MeteoGalicia \cite{bib_meteogalicia}) consisted until now of a global description of the short-term meteorological trend (Fig. \ref{forecastapril}). This service has been recently improved in order to provide visitors with symbolic forecasts for each of the 315 municipalities in Galicia, thus improving its quality and allowing users to obtain more precise information about specific locations of the Galician geography. 

\begin{figure}[!t]
\centering
\includegraphics[width=0.9\columnwidth]{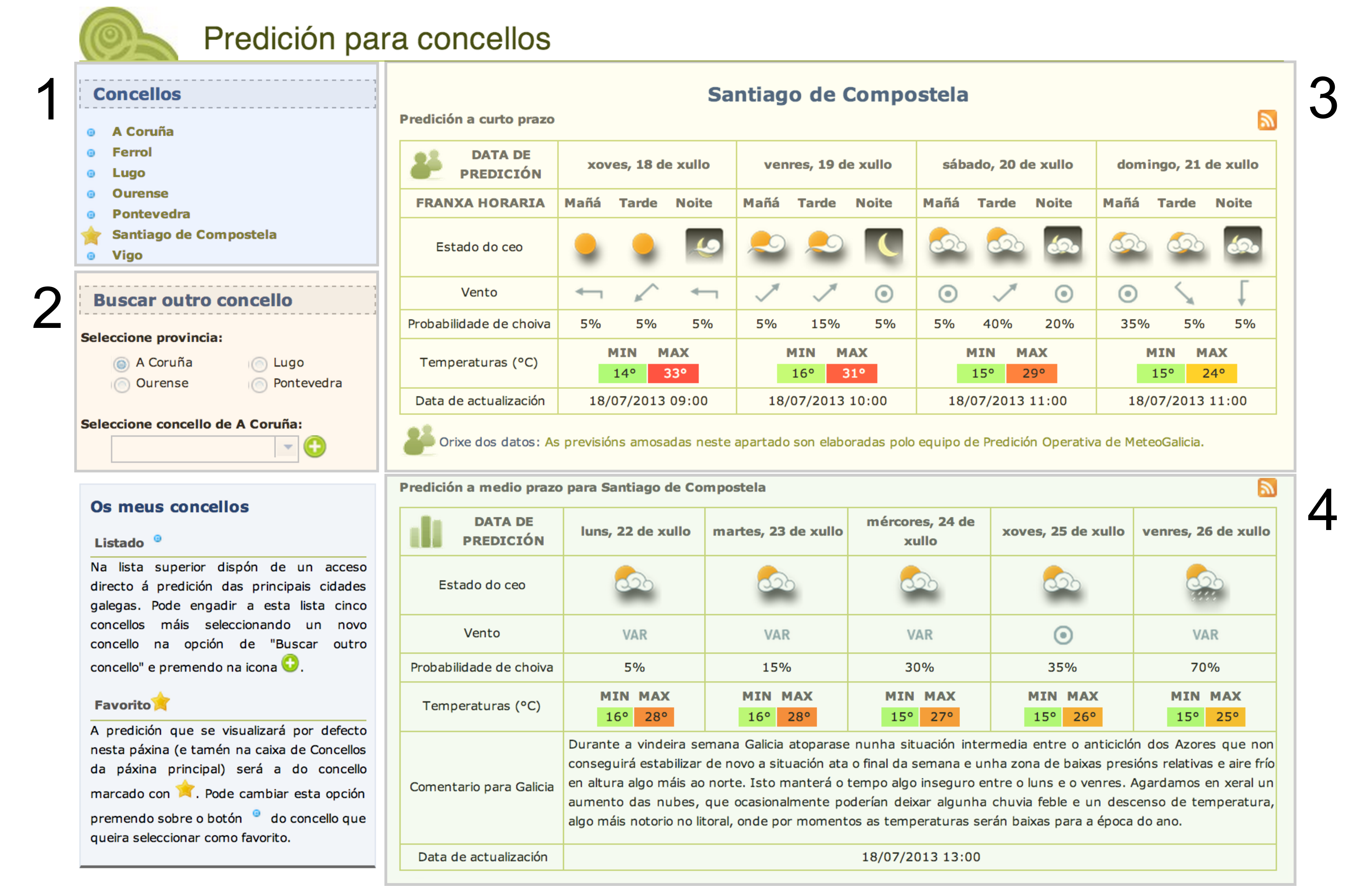}
\caption{Short-term and mid-term municipality forecast web application for Galicia \cite{bib_meteogalicia}.} 
\label{webforecast}
\end{figure}

Figure \ref{webforecast} shows the current web application for consulting municipality forecasts \cite{bib_meteogalicia}, which has been graphically divided in blocks for an easier explanation. Block 1 contains a shortcut list to the seven most important municipalities in Galicia, which allows a direct access to their forecast data (the user can select a favorite municipality, which is loaded by default in posterior visits). Block 2 allows the user to search for the rest of the municipalities, which are grouped according to the Galician province they belong to. It also allows to add to the shortcut list in Block 1 the selected municipality. The short-term forecast is shown in Block 3, which offers symbolic data for wind and sky state and numeric data for temperatures for four days, including morning, afternoon and night each day. Block 4 shows the mid-term forecast for several days and includes a global comment about the weather in Galicia in general, which consequently remains the same for every municipality.

This increase in the quantity of available numerical-symbolic data has a main downside, which resides in the lack of natural language forecasts which describe this set of data. This issue makes forecasts harder to understand, since users need to look at every symbol and detect which phenomena are relevant and when they will occur, whereas natural language descriptions directly provide all this information. In the case of a mid-term forecast, its uncertainty allows the inclusion of a global description, which is written by a meteorologist. However, for short-term forecasts, which are much more accurate, the meteorological diversity causes that several meteorological phenomena may occur at the same time in different areas. Thus, to issue daily textual forecasts upon 315 municipalities is not feasible.

In order to address this issue we have developed an application which, from short-term data, generates linguistic descriptions which highlight meteorological phenomena considered important by an expert meteorologist. The style and contents of the natural language linguistic descriptions for each location are similar to the general one presented in Fig. \ref{forecastapril}.

\section{Application description}

\begin{figure}[!b]
\centering
\includegraphics[width=0.6\columnwidth]{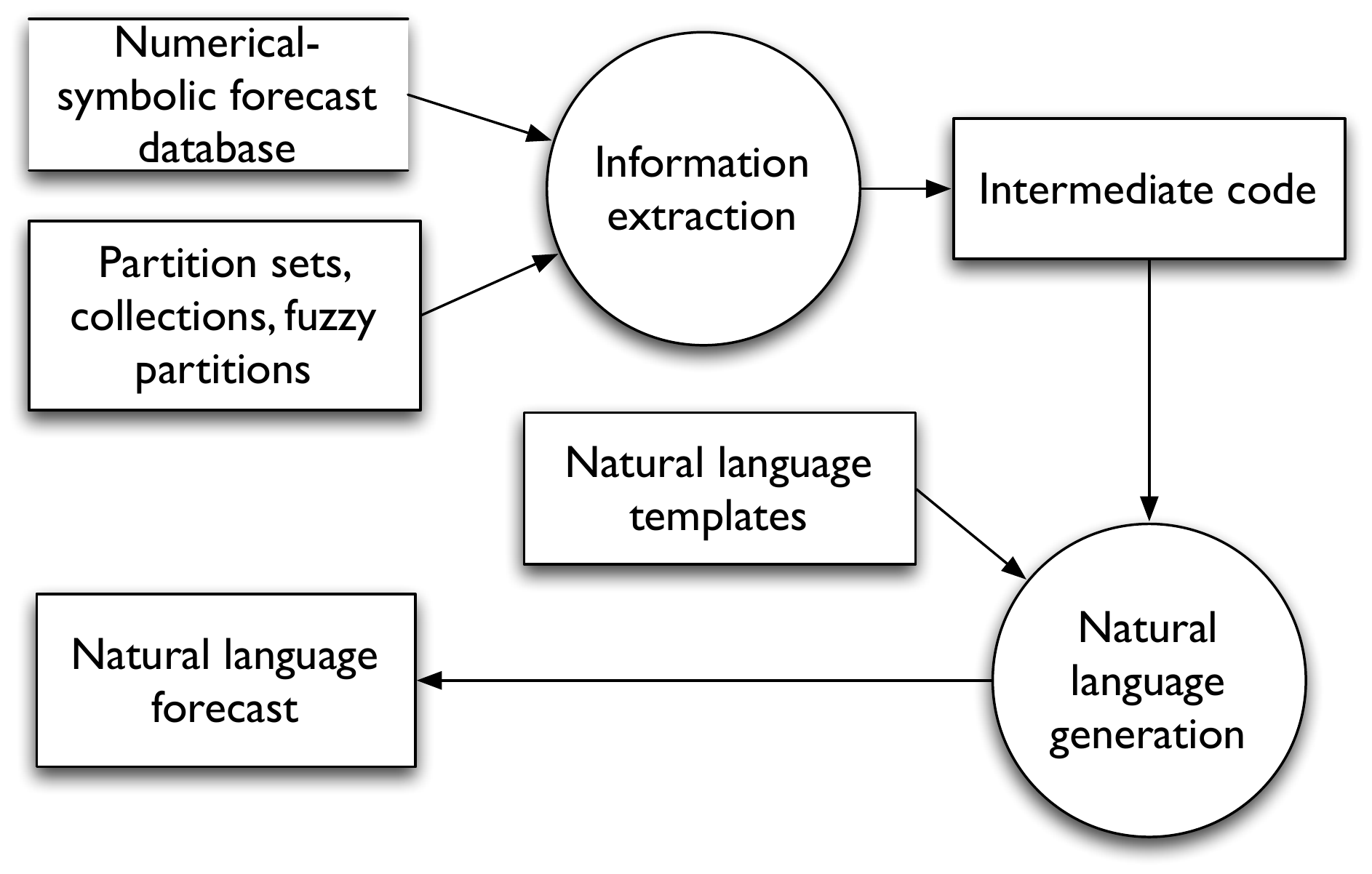}
\caption{General schema of the application architecture.} 
\label{methodschema}
\end{figure}

The solution we have devised employs numerical-symbolic forecast data and additional expert information to generate the final output textual weather forecasts in two separate tasks. The first task converts the numerical-symbolic input data into linguistic descriptions (encoded in an intermediate language). These descriptions are created through a computational method which abstracts data values into linguistic labels dealing with uncertainty and temporal references. In the second stage, a NLG system translates the intermediate codes into a natural language forecast for one of the available final output natural languages, which is ready for human consumption. A general schema of this process is shown in Fig. \ref{methodschema}.

\subsection{Input weather forecast data characterization}
\begin{figure}[!b]
\centering
\includegraphics[width=0.8\columnwidth]{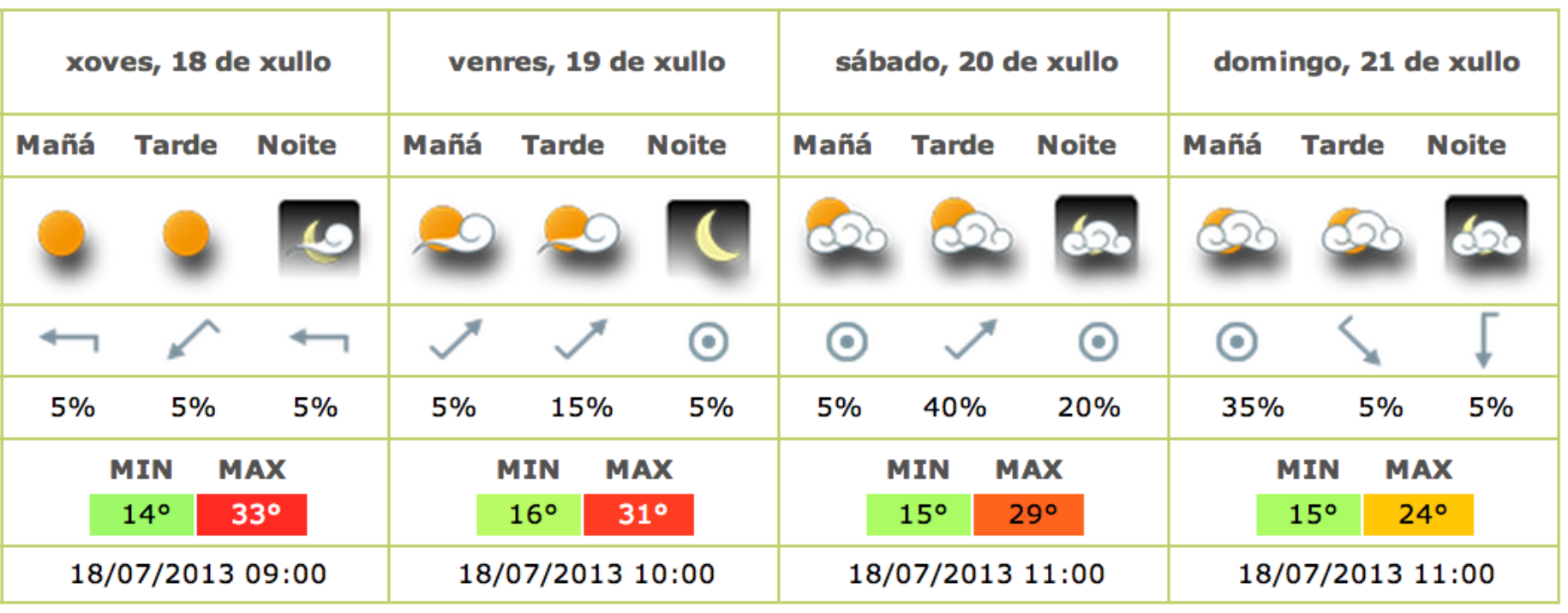}
\caption{Real example of a data source for a given location used in the generation of the automatic weather forecasts.} 
\label{short}
\end{figure}

MeteoGalicia's database offers a dataset which covers all the 315 Galician municipalities and includes forecast data associated to several items in a four-day temporal window. This data is heterogeneous in its nature and includes values in degrees Celsius and weather symbols represented by codes. For instance, the meteorologists have characterized the sky state phenomena as 21 numerical codes (values in the interval [101,121]) and the wind phenomena as 34 numerical codes (values associated to a given intensity and direction in the interval [299,332]). These numerical codes are used to display graphical symbols in the forecast website. Figure \ref{short} shows an example of a real short-term forecast data series.

Formally, each municipality $M$ has an associated forecast data series set $\allowbreak FD_M = \{SS_M, \allowbreak W_M, TMAX_M, TMIN_M\}$ , which includes data series for the input variables considered: sky state ($SS_M$), wind ($W_M$) and maximum ($TMAX_M$) and minimum ($TMIN_M$) temperatures. For clarity reasons, without loss of generality, we will consider a single municipality data series in the explanations that follow ($FD_M = FD$). Each data series element in $FD$ is characterized in what follows:
\begin{itemize}
		\item \textbf{Sky state ($SS$)}. It provides three numerical codes per day (morning, afternoon, night) about two meteorological variables of interest, namely \underline{cloud coverage} and \underline{precipitation}. From a formal point of view, $SS = \{ss_1, \ldots, ss_i, \ldots, ss_{12}\}$, where $ss_i \in [101,121] \forall ss_i \in SS$. Each code in the interval $[101,121]$ has a specific sky state meaning (for example, 111 means ``covered with rain'').
		\item \textbf{Wind  ($W$)}. It provides three numerical codes per day about the wind intensity and direction. $W = \{w_1,...,w_i,...,w_{12}\}$, where $w_i \in [299,332] \forall w_i \in W$. Each code in the interval $[299,332]$ has an associated wind direction and intensity (for instance, 317 means ``strong wind from the North'').
		\item \textbf{Temperature} ($TMAX$ and $TMIN$). Maximum and minimum forecasted temperatures are given in degrees Celsius with a resolution of 1 degree and one value per day:
			\begin{itemize}
				\item $TMAX = \{tmax_1,tmax_2,tmax_3,tmax_4\}$, where $tmax_i \in [-60^{\circ}C, 60^{\circ}C] \forall tmax_i \in TMAX$.
				\item $TMIN = \{tmin_1,tmin_2,tmin_3,tmin_4\}$, where $tmin_i \in [-60^{\circ} C, 60^{\circ}C] \forall tmin_i \in TMIN$.
			\end{itemize} 
	\end{itemize}
	
For each forecast data series $FD$, our application obtains linguistic descriptions about seven forecast variables, namely cloud coverage, precipitation, wind, maximum and minimum temperature variation and maximum and minimum temperature climatic behavior \footnote{It measures the difference between the forecasted temperatures and the temperature climatic mean, defined as the average for the previous 30 years in a given month.}. For this, we have devised a computational method divided in several linguistic description generation operators.

\subsection{First stage: Linguistic description generation method}
\begin{figure}[!b]
\centering
\includegraphics[width=0.8\columnwidth]{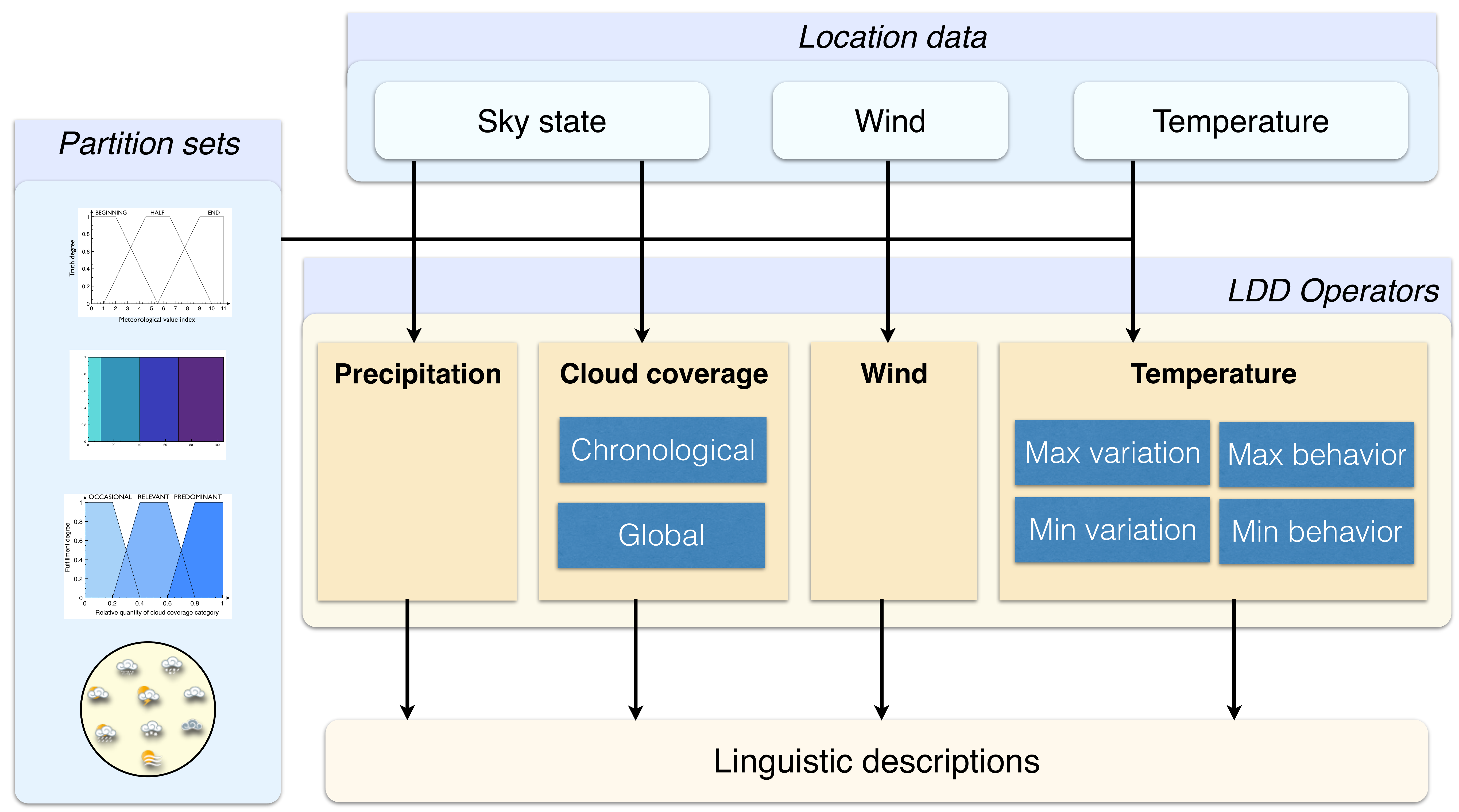}
\caption{Global schema of the linguistic description generation method.} 
\label{firstphase}
\end{figure}

The first stage of our application obtains a linguistic description for every variable, which consists in sets of linguistic labels and temporal references which contain the relevant information extracted from the raw data. This process, as it can be seen in Fig. \ref{firstphase}, consists of providing each linguistic description operator with its corresponding data and expert knowledge (in the form of crisp and fuzzy partition sets and numeric categories) in order to generate the intermediate linguistic descriptions. Each operator is formally described in what follows.

\subsubsection{Cloud coverage fuzzy operators}
\begin{figure}[!t]
\centering
\includegraphics[width=0.9\columnwidth]{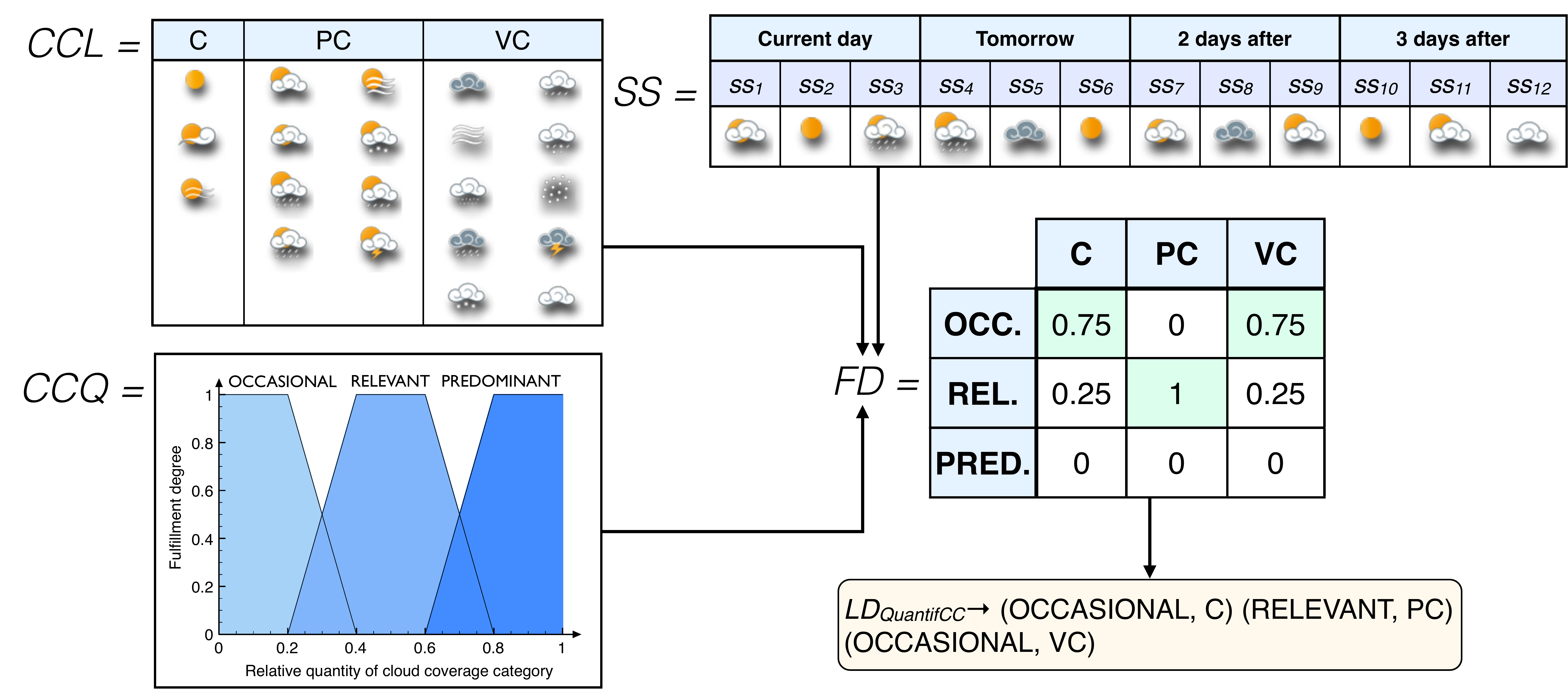}
\caption{Global quantification description fuzzy operator definitions and process example.} 
\label{fuzzyquant}
\end{figure}

Two different fuzzy operators are used in the linguistic description generation of the cloud coverage variable. The first one provides a chronological description, while the second one provides a short-term global description when the previous description is not appropriate.
	\begin{enumerate}
		\item \underline{Chronological description fuzzy operator}.
		\begin{itemize}
			\item \textbf{Input}:
				\begin{itemize}
					\item Sky state data series $SS = \{ss_1, \ldots, ss_i, \ldots, ss_{12}\}$.
					\item A temporal fuzzy linguistic partition $CCT = \{cct_1, \ldots, cct_j, \ldots, cct_n\}$, where each temporal linguistic term $cct_j$ has an associated fuzzy membership function $\mu_{cct_j} \colon \mathbb{N} \rightarrow [0,1]$. For our application, $CCT = \{BEGINNING, HALF, END \}$ (Fig. \ref{fuzzytime}).
					\item A cloud coverage linguistic variable, defined as a set of cloud coverage categories $CCL = \{ccl_1,\ldots,ccl_k,\ldots,ccl_m\}$. Each linguistic term $ccl_k \in CCL$ has an associated crisp membership function $\mu_{ccl_k} \colon \mathbb{N} \rightarrow \{0,1 \}$, defined as:
					\begin{equation}
					\mu_{ccl_k}(ss_i) =
					\begin{cases}
					   1 & \text{if } ss_i \in ccl_k \\
					   0  & \text{otherwise}
					\end{cases}
					\label{eq1}
					\end{equation}
				\end{itemize}
			In our application, $CCL = \{C,PC,VC \}$ (``clear'', ``partly cloudy'', ``very cloudy''), as shown in Fig. \ref{fuzzytime}.
			\item \textbf{Procedure}. This operator provides the most appropriate cloud coverage linguistic term $ccl_k$ for each temporal subdivision $cct_j$. A relevance degree is calculated for each pair of cloud coverage and temporal labels and the label pairs with the highest degree are then selected (one per temporal label):
			\begin{itemize}
				\item Relevance degree matrix $RD$, where each value $RD_{j,k}$ determines the importance a cloud coverage linguistic term $ccl_k$ has within a temporal sub period $cct_j$: $RD_{j,k} = \sum\limits_{i=1}^{|SS|} \mu_{ccl_k}(ss_i)*\mu_{cct_j}(i)$
				\item Set of the most appropriate cloud coverage label for each temporal label, ordered by the temporal partition index $j$: $CCTL = \{ (cct_j ,ccl_k) | RD_{j,k} = max(RD_j)\}$
			\end{itemize}
		\item \textbf{Output}. A chronological cloud coverage linguistic description as an intermediate code characterized by the following concatenation: \[LD_{ChronoCC} \rightarrow (cct_1,ccl_k) \ldots (cct_n, ccl_k)\]
		\end{itemize}
		Figure \ref{fuzzytime} shows the definitions of both linguistic variables for our application and an example of the chronological cloud coverage linguistic description process. This description is provided only if the following experimental condition is fulfilled: $\forall (cct_j, ccl_k) \in CCTL, RD_{j,k} \geq 3$. This condition ensures that every $cct_j$ has an associated predominant cloud coverage type $ccl_k$, while maintaining tolerance to the appearance of other cloud coverage categories in $SS$. Otherwise, the linguistic description generated by the second operator is provided.
		
\begin{figure}[!t]
\centering
\includegraphics[width=0.9\columnwidth]{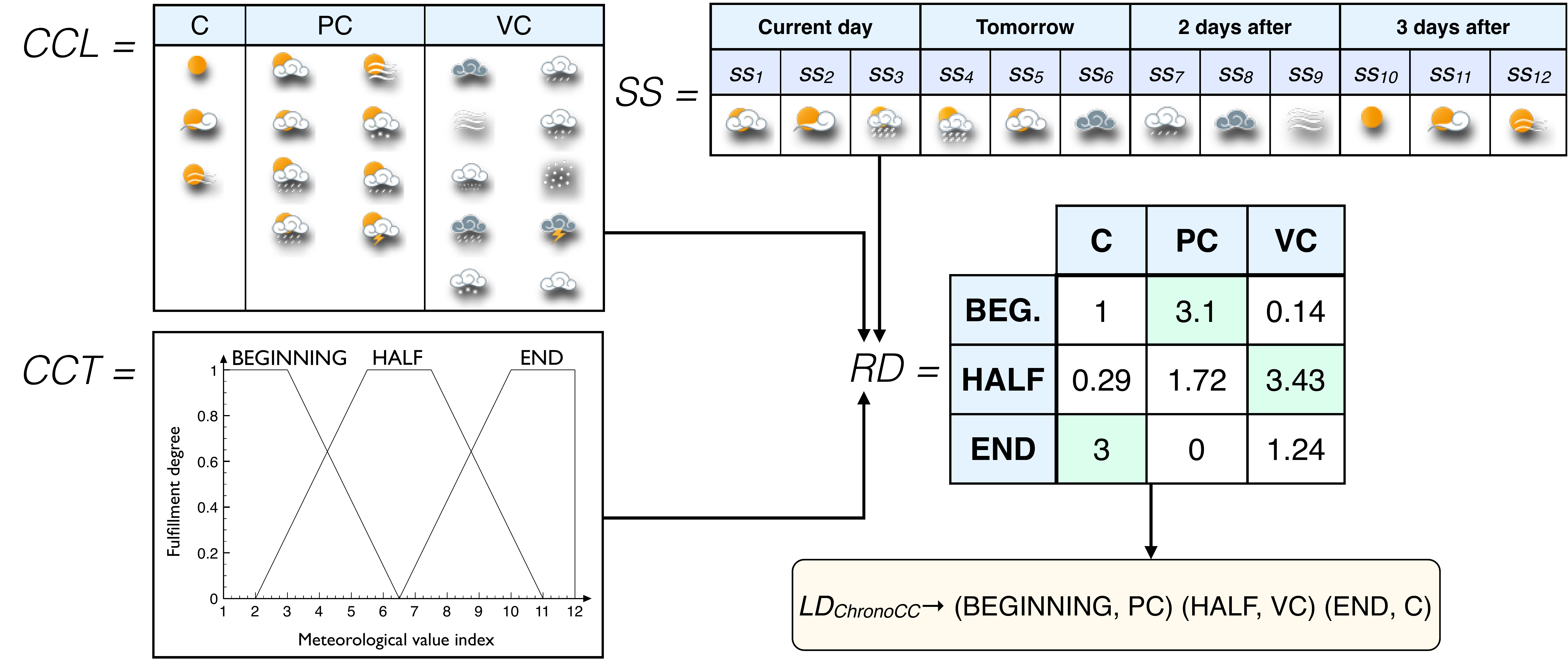}
\caption{Chronological description fuzzy operator definitions and process example.} 
\label{fuzzytime}
\end{figure}

\item \underline{Global quantification description fuzzy operator}. This operator provides a global description of the cloud coverage state for the whole short-term period.
		\begin{itemize}
		\item \textbf{Input}:
					\begin{itemize}
						\item Sky state data series $SS = \{ss_1, \ldots, ss_i, \ldots, ss_{12}\}$.
						\item A cloud coverage predominance linguistic label $CCQ = \{ccq_1, \ldots, ccq_j, \ldots, ccq_n\}$, where each linguistic term $ccq_j$ has an associated fuzzy quantifier $\mu_{ccq_j} \colon [0,1] \rightarrow [0,1]$. In our case, $CCQ = \{OCCASIONAL, RELEVANT,$ $PREDOMINANT\}$ (Fig. \ref{fuzzyquant}).
						\item A cloud coverage linguistic variable $CCL$, as defined in the previous operator.
					\end{itemize}
		\item \textbf{Procedure}. This operator quantifies the occurrence of the different cloud coverage categories $ccl_k$ using Zadeh's quantification model \cite{bib_zadehquantif}:
				\begin{itemize}
					\item Fuzzy fulfillment degree matrix $FD$, where $FD_{j,k} = \mu_{ccq_j}(\sum\limits_{i=1}^{|SS|}\frac{ \mu_{ccl_k}(ss_i)}{|SS|})$
					\item Set of cloud coverage label and quantifier label pairs with the highest fulfillment degree: $CCQL = \{(ccq_j,ccl_k) | FD_{j,k} = 
 \operatorname*{max}_l FD_{l,k}\}$, where $j$ is minimum.
				\end{itemize}
			\item \textbf{Output}. A cloud coverage linguistic description as an intermediate code characterized by the following concatenation: \[LD_{QuantifCC} \rightarrow (ccq_j, ccl_1) \ldots (ccq_j, ccl_m)\]
		\end{itemize}
		 Figure \ref{fuzzyquant} shows the definition of the fuzzy quantifiers $\mu_{ccq_j}$ and an example of this linguistic description process.
	\end{enumerate}
	
\subsubsection{Precipitation episode extractor operator}
This operator extracts precipitation episodes from the sky state values. These periods are classified according to the kind of precipitations detected:

\begin{itemize}
\item \textbf{Input}:
	\begin{itemize}
		\item Sky state data series $SS = \{ss_1, \ldots, ss_i, \ldots, ss_{12}\}$.
		\item A precipitation linguistic variable, defined as a set of precipitation categories $PV = \{pv_1,\ldots,pv_j,\ldots,pv_n\}$, where each linguistic term $pv_j$ has an associated crisp membership function $\mu_{pv_j} \colon \mathbb{N} \rightarrow \{0,1 \}$, where $\mu_{pv_j}$ is defined identically as $\mu_{ccl_k}$ in expression (\ref{eq1}).
	\end{itemize}
\item \textbf{Procedure}. This operator extracts an ordered set of precipitation episodes $PE = \{pe_1, \allowbreak \ldots, pe_k, \ldots ,pe_m \}$, where each episode is characterized as $pe_k = \{START, END, \allowbreak LABELS\}$. The algorithm in Fig. \ref{precipprocedure} describes how the precipitation operator extracts the relevant episodes from $SS$:

\begin{figure}[!t]
\centering
\begin{algorithmic}
\small
\Procedure{PrecipitationEpisodeExtractor}{$SS$,$PV$}
\State $PE \leftarrow \{\}$
\State $pe_k \gets \O$
\While{$i < |SS|$}
	\State $active\_period \gets False$
	\ForAll{$pv_j \in PV$}
		\If{$\mu_{pv_j}(ss_i) = 1$}
			\If{$ pe_k \not= \O $}
				\State $pe_k.LABELS \gets pe_k.LABELS \cup pv_j$
			\Else
				\State $pe_k \gets \{START, END, LABELS\}$
				\State $pe_k.START \gets i$
				\State $pe_k.LABELS \gets \{pv_j\}$
				\State $PE \gets PE \cup pe_k$
			\EndIf
			\State $active\_period \gets True$
			\State break
		\EndIf
	\EndFor
	\If{$\neg active\_period \And pe_k \not= \O$}
		\State $pe_k.END \gets i-1$
		\State $pe_k \gets \O$
	\EndIf
	\State $i \gets i + 1$
\EndWhile
\If{$ active\_period \And pe_k \not= \O$}
		\State $pe_k.END \gets |SS|$
\EndIf
\State \textbf{return} $PE$
\EndProcedure
\end{algorithmic}
\caption{Precipitation episode extractor procedure.} 
\label{precipprocedure}
\end{figure}

	\item \textbf{Output}. A precipitation linguistic description for each precipitation episode $pe_k$ as an intermediate code characterized by the following concatenation of terms: \[LD_{Precipitation_k} \rightarrow START_k\: END_k\: LABELS_k\]
\end{itemize}

\begin{figure}[!b]
\centering
\includegraphics[width=0.8\columnwidth]{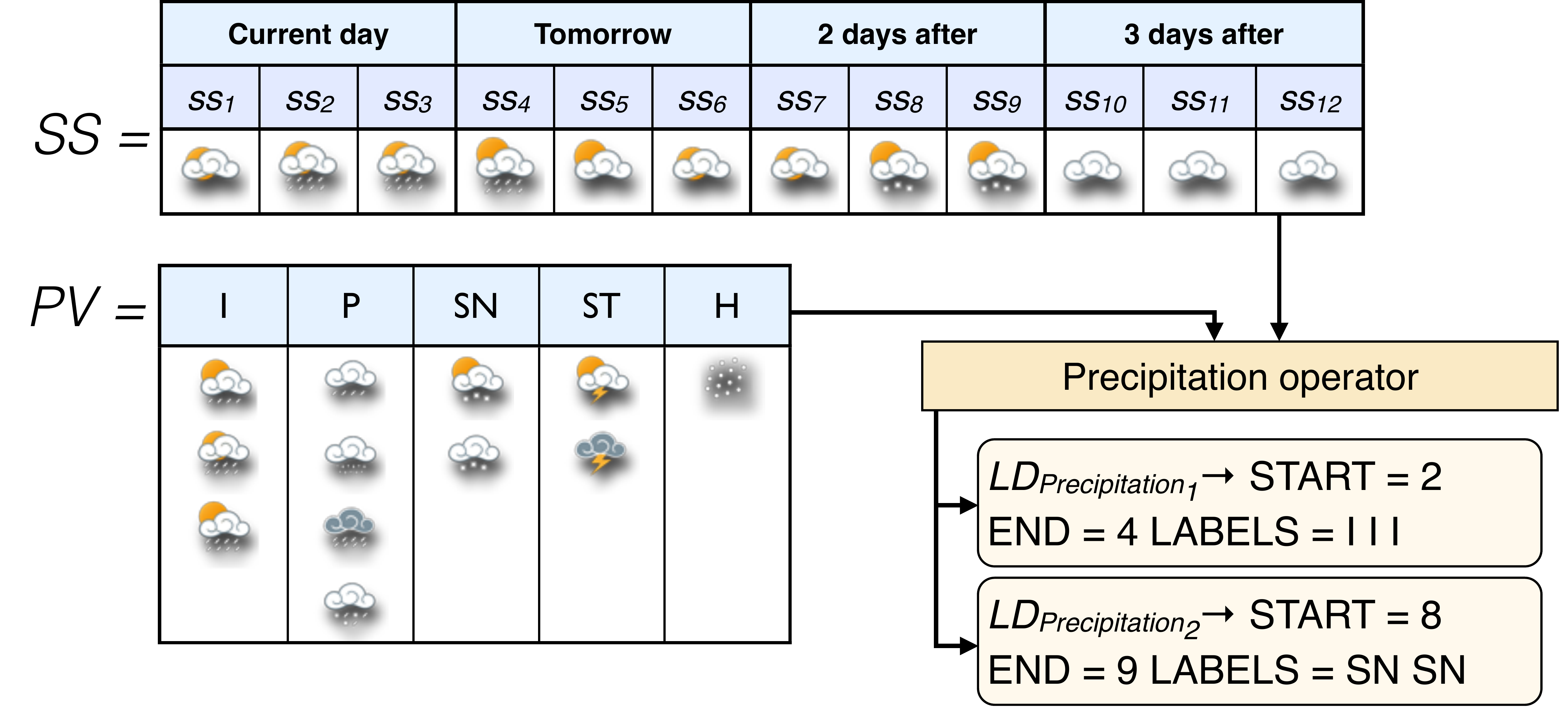}
\caption{Schema of the precipitation operator method with the current meteorological phenomena categories for precipitation and its associated labels.} 
\label{precipitation}
\end{figure}

In this case, $PL = \{I,P,SN,ST,H \}$ (``intermittent'',``persistent'',``snow'',``storm'',``hail'') is defined for precipitation (although ``intermittent'' and ``persistent'' are not explicitly included in the final natural language forecasts, as required by the meteorologists). Figure \ref{precipitation} shows the definition of $PL$ and provides a graphical example of the precipitation linguistic description generation process.

\subsubsection{Wind operator}
It follows a similar strategy to the precipitation operator, although in this case it does not convert the original values into labels.
\begin{itemize}
\item \textbf{Input}:
	\begin{itemize}
		\item Wind data series $W = \{w_1, \ldots, w_i, \ldots, w_{12}\}$.
		\item A numeric interval $AW = [aw_a,aw_b] | AW \subset [299,332]$ (as indicated in Section III-A), which specifies the relevant wind values to be extracted by the operator. In our application, $AW = [317,332]$. This interval corresponds to strong and very strong winds, which are the only relevant wind conditions to be included in the descriptions according to the meteorologists.
	\end{itemize}
\item \textbf{Procedure}.  This operator extracts an ordered set of wind episodes $WE = \{we_1, \allowbreak \ldots, we_k, \ldots ,we_m \}$, where each episode is characterized as $we_k = \{START_k, END_k,\allowbreak SYMBOLS_k\}$. The algorithm in Fig. \ref{windprocedure} describes how the wind operator extracts the relevant episodes from $W$.

\begin{figure}[!b]
\centering
\begin{algorithmic}
\small
\Procedure{WindEpisodeExtractor}{$W,AW$}
\State $WE \leftarrow \{\}$
\State $we_k \gets \O$
\While{$i < |W|$}
	\State $active\_period \gets False$
	\If{$ w_i \in AW$}
		\If{$ we_k \not= \O $}
			\State $we_k.SYMBOLS \gets we_k.SYMBOLS \cup w_i$
		\Else
			\State $we_k \gets \{START, END, LABELS\}$
			\State $we_k.START \gets i$
			\State $we_k.SYMBOLS \gets \{w_i\}$
			\State $WE \gets WE \cup we_k$
		\EndIf
		\State $active\_period \gets True$
	\EndIf
	\If{$\neg active\_period \And we_k \not= \O$}
		\State $we_k.END \gets i-1$
		\State $we_k \gets \O$
	\EndIf
	\State $i \gets i + 1$
\EndWhile
\If{$ active\_period \And we_k \not= \O$}
		\State $we_k.END \gets |W|$
\EndIf
\State \textbf{return} $WE$
\EndProcedure
\end{algorithmic}
\caption{Wind episode extractor algorithm.} 
\label{windprocedure}
\end{figure}

	\item \textbf{Output}. A wind linguistic description for each wind episode $we_j$ as an intermediate code characterized by the following concatenation: $LD_{Wind_k} \rightarrow$ $START_k$ $END_k$ $SYMBOLS_k$. For example, if there is a period of strong wind within $W$, we could obtain a linguistic description such as ``START=2 END=4 LABELS=322,322,322'', meaning ``from tonight ($i = 2$) until tomorrow afternoon ($i=4$) there will be strong wind from the southwest ($w_i= 322$)''.
\end{itemize}

\subsubsection{Temperature operator}
This operator generates a linguistic description which reflects the temperature trend for the 4-day period and also obtains information about the climatic behavior of the forecasted temperatures. Thus, four variables are considered: maximum and minimum temperature variations and maximum and minimum climatic behavior.
\begin{itemize}
\item \textbf{Input}:
	\begin{itemize}
		\item Maximum temperature data series $TMAX = \{tmax_1, tmax_2, tmax_3, tmax_4\}$.
		\item Minimum temperature data series $TMIN = \{tmin_1,tmin_2, tmin_3,tmin_4\}$.
		\item A temperature variation linguistic variable, defined as $TV = \{tv_1,\ldots,tv_j,\ldots,tv_n\}$, where each linguistic term $tv_j \in TV$ has an associated crisp membership function $\mu_{tv_j} \colon \mathbb{R} \rightarrow \{0,1 \}$. In our application, $TV = \{ED,ND,MD,SD,WC,SI,MI,NI,EI \}$ (``extreme decrease'', ``notable decrease'', ``moderate decrease'', ``slight decrease'', ``without changes'', ``slight increase'', ..., ``extreme increase'').
		\item A temperature climatic behavior linguistic variable, defined as $TC = \{tc_1,\ldots,tc_j,\allowbreak \ldots,tc_n\}$, where each linguistic term $tc_j \in TC$ has an associated crisp membership function $\mu_{tc_j} \colon \mathbb{R} \rightarrow \{0,1 \}$. In our case, $TC = \{VL,L,N,H,VH\}$ (``very low'', ``low'', ``normal'', ``high'', ``very high''). 
	\end{itemize}
\item \textbf{Procedure}. This operator provides the linguistic terms with the highest membership degree from $TV$ and $TC$ for the four temperature variables considered:
	\begin{itemize}
	\item Temperature variation: for maxima $TMAXV = tv_j | \mu_{tv_j}(tmax_{|TMAX|} - tmax_1) = 1$, and minima $TMINV = tv_j | \mu_{tv_j}(tmin_{|TMIN|} - tmin_1) = 1$.
	\item Temperature climatic behavior: for maxima $TMAXC = tc_j | \mu_{tc_j}(\sum\limits_{i=1}^{|TMAX|} \frac{tmax_i}{|TMAX|}) = 1$, and minima $TMINC = tc_j | \mu_{tc_j}(\sum\limits_{i=1}^{|TMIN|} \frac{tmin_i}{|TMIN|}) = 1$.
	\end{itemize}
	\item \textbf{Output}. A temperature linguistic description as an intermediate code characterized by the following term concatenation: \[LD_{Temperature} \rightarrow TMINC \;TMAXC \;TMINV\] \[TMAXV\]
\end{itemize}

\begin{figure}[!b]
\centering
\includegraphics[width=0.7\columnwidth]{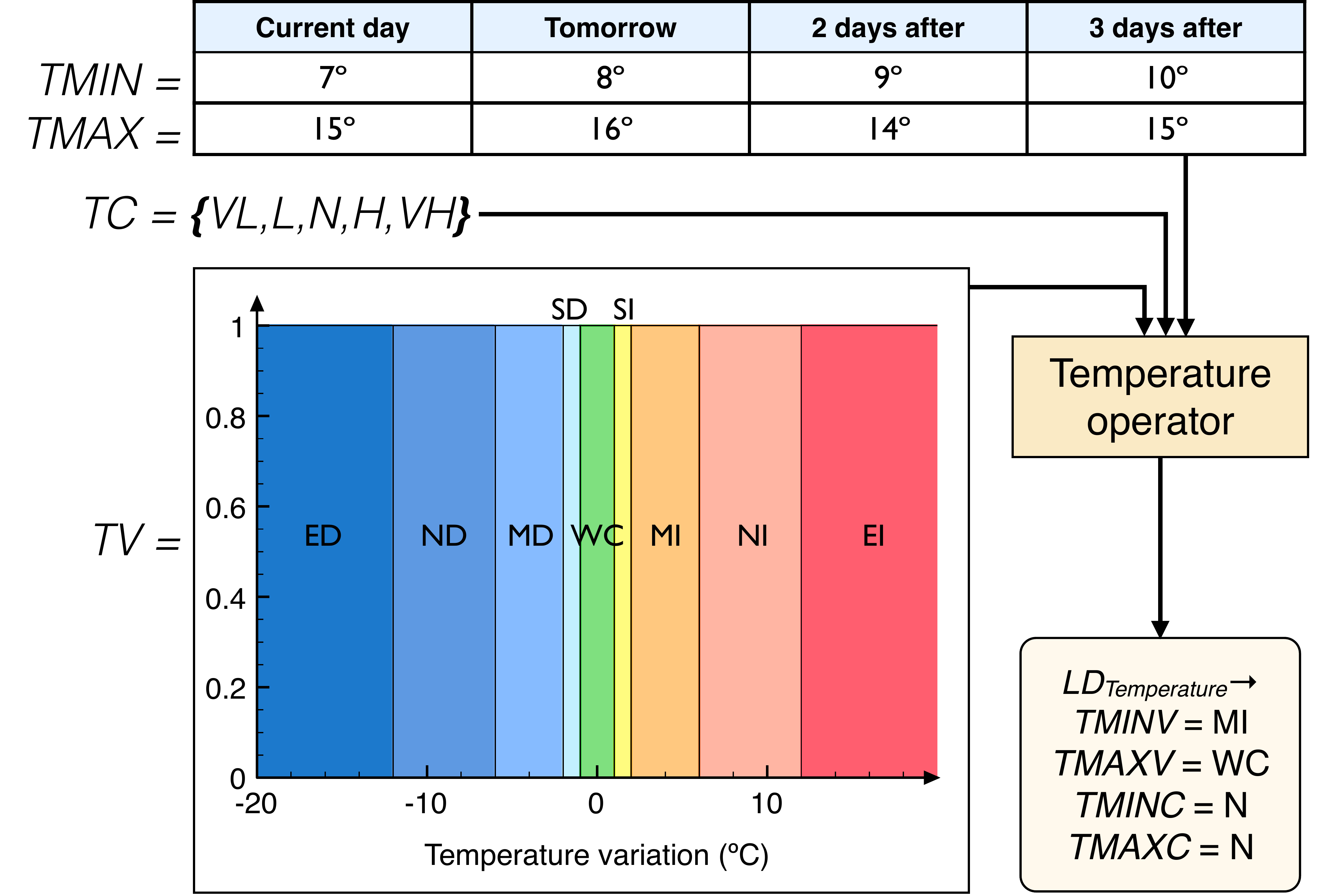}
\caption{Schema of the temperature operator, with the current definition of the temperature variation partition and its associated labels.}
\label{temperature}
\end{figure}

The definition of $TV$ and a graphical example of the temperature operator are shown in Figure \ref{temperature}. As for $TC$, its associated crisp membership functions $\mu_{tc_j}$ are not shown in this example, since they vary for each municipality.

\subsection{Second stage: Natural language generation}
The natural language generation (NLG) stage of this application consists of a domain-specific system which, following standard NLG techniques, has also been divided into different modules for each variable, so that changes in one of them do not affect the rest of the system. From a global perspective, each of these modules receives the intermediate linguistic description generated by their corresponding operator, parses it and generates the final textual forecast for its associated variable.

If we delve deeper into the natural language generation stage structure, the complexity of the final natural language descriptions is a factor which has determined the design and implementation approach we have followed. This includes evaluation criteria applicable to linguistic descriptions \cite{bib_felixisda} such as the description length, but also NLG systems design methodologies as in \cite{bib_building_nlg} and \cite{bib_building_nlgart}.

Thus, since the quantity of information in the descriptions is variable and the diversity of situations for each variable to be included ranges from simple to more complex, we have adopted two different NLG solutions. On one hand, we have defined templates in structured text files which contain generic natural language sentences for the simpler variables (cloud coverage, temperatures and wind). On the other hand, we have designed and implemented the generation of natural language sentences for precipitation inspired by standard NLG methodologies \cite{bib_building_nlg}, \cite{bib_building_nlgart}.

\subsubsection{Template-based NLG approach}

This approach has been devised as a solution for variables whose corresponding natural language sentences have rather static structure and length, such as temperatures or cloud coverage. For example, a textual forecast for temperatures usually includes information about variation of maxima and minima and their climate behavior, and the only elements that differ from one forecast to another are the labels assigned to the variations and the behavior, whereas the syntactic structure and length of the forecasts remain the same.

\begin{figure}[!b]
\centering
\includegraphics[width=1\columnwidth]{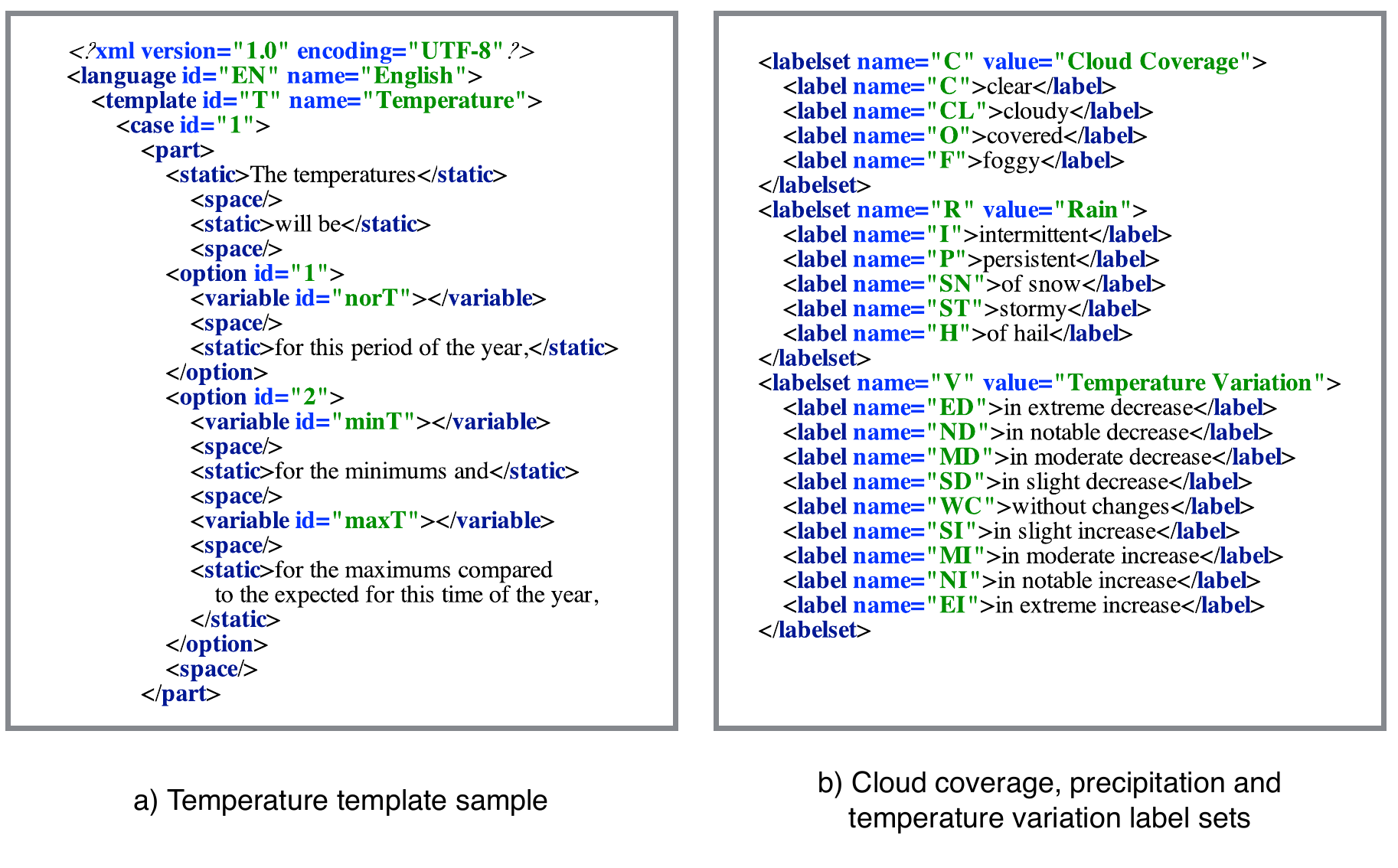}
\caption{Temperature template sample and label sets from the English language template document.} 
\label{xml}
\end{figure}

\begin{figure}[!t]
\centering
\includegraphics[width=0.9\columnwidth]{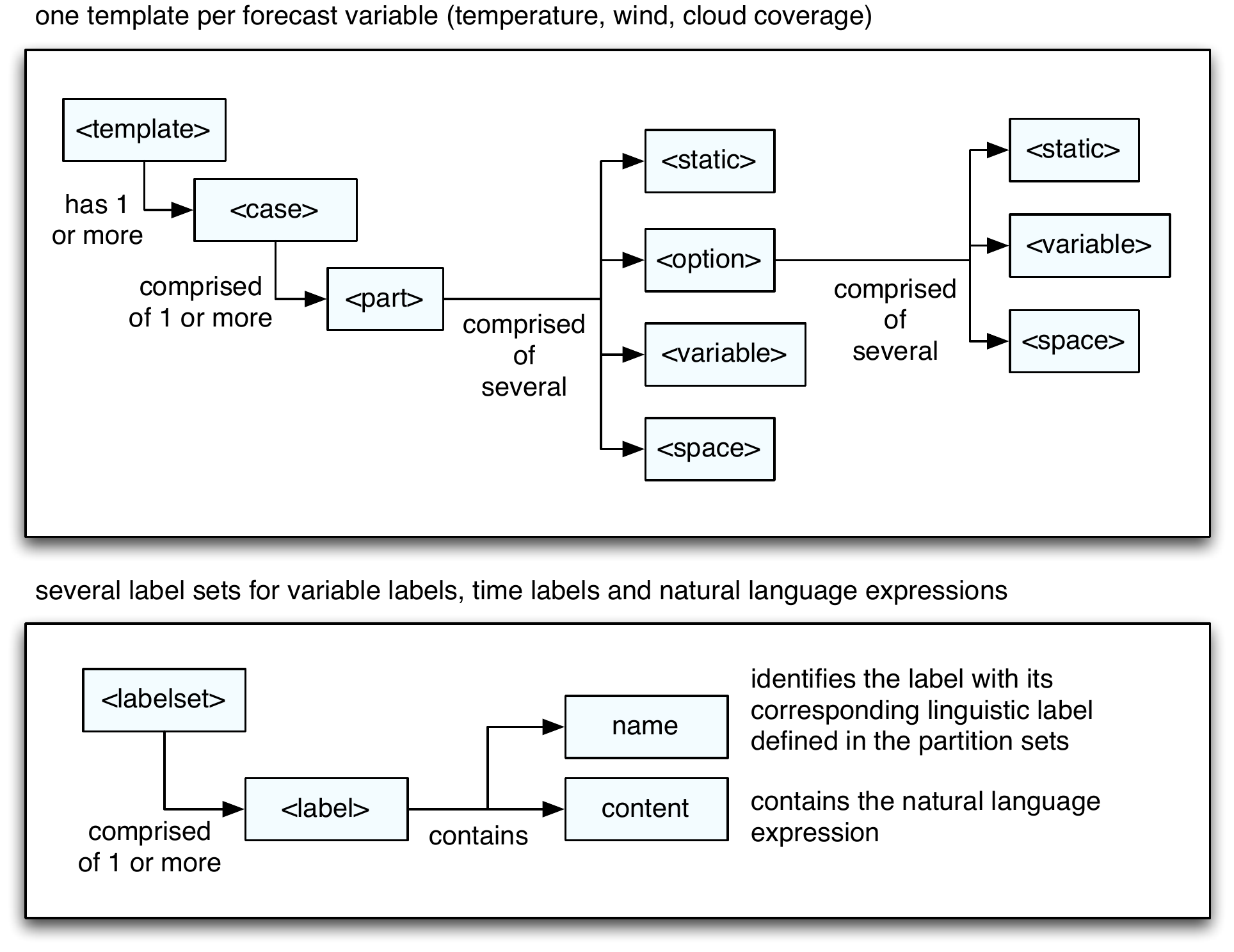}
\caption{Schema of the structure of a NLG template file, which contains generic sentences and label sets.} 
\label{nlg_structures}
\end{figure}

In this context, structured text files, such as XML, allow to model and build templates of natural language sentences, where static text can be mixed with other elements, such as variables or optional texts within a sentence. We have taken advantage of this flexibility by designing templates for temperature, cloud coverage and wind variables. These templates are included in a document which also contains natural language label sets for variables, time expressions or other kind of language-dependent text resources. Figure \ref{xml} shows parts of a template document (in this case for English language), whose structure (Fig. \ref{nlg_structures}) is comprised of the following elements:

\begin{itemize}
\item \textbf{Variable templates}, which include the generic natural language forecast structures for several variables, such as cloud coverage or temperature.
\item \textbf{Label sets}, which contain the natural language vocabulary and expressions used to fill in the variable elements. They are the natural language equivalent to the crisp and fuzzy partition sets used in the linguistic description extraction stage. For example, in Fig. \ref{temperature} the temperature variation labels in $TV$ correspond to the label identifiers in the temperature variation label set in Fig. \ref{xml}.
\end{itemize}

The template documents for the supported languages are loaded into structured objects within the application. Once the intermediate codes for the NLG template-based variables have been obtained, each NLG module (one per meteorological variable) parses its corresponding code and executes expert rules incorporated into the implementation code, so that according to certain detectable events in the intermediate language, different cases and options can be selected. Then, the template variables are filled with the natural language labels which correspond to the linguistic labels found in the intermediate code. Finally, the NLG template structures are translated into a natural language forecast text through the concatenation of the text values of each of their elements.

\subsubsection{Precipitation NLG approach}
The previous NLG approach is not suitable for variables such as precipitation, where several episodes can occur within a forecast term. This can lead to the generation of several natural language sentences which, although may reflect faithfully the meteorological data, are repetitive and tedious to read. Since the purpose of building linguistic descriptions in natural language is to provide users with textual information which should be easy to read and to understand, another NLG approach is required in order to achieve this goal.

Based on the concepts of a NLG system architecture described in \cite{bib_building_nlg} and \cite{bib_building_nlgart}, we have designed and developed a NLG module for precipitation which addresses redundancy or length excess in the obtained descriptions.

\begin{figure}[!t]
\centering
\includegraphics[width=0.7\columnwidth]{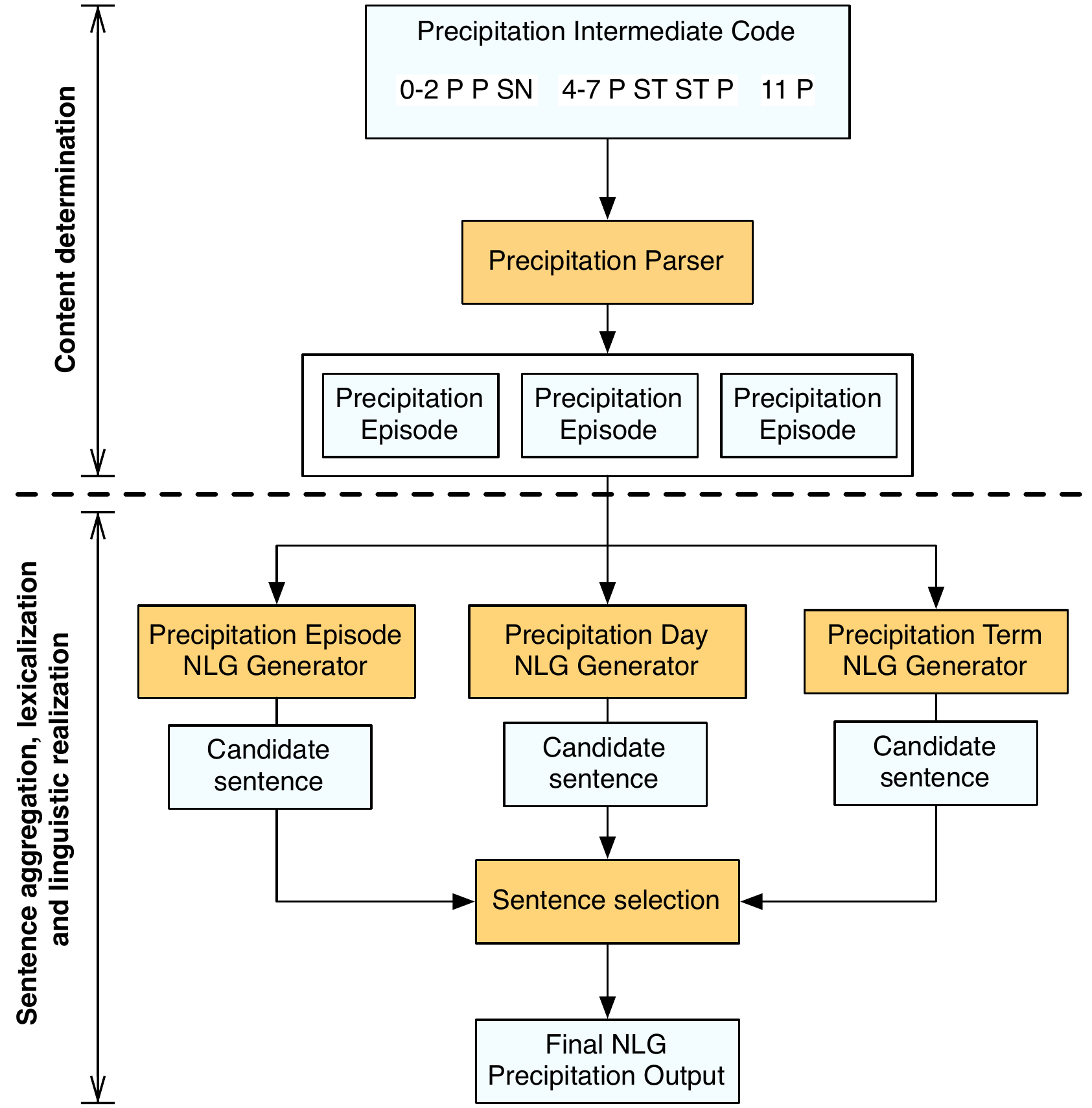}
\caption{Schema of the NLG approach for the precipitation variable.} 
\label{precnlg}
\end{figure}

In \cite{bib_building_nlgart}, a NLG system is depicted as a six stage task, where one subtask is performed per stage. However, some of these subtasks may be merged or might not even be necessary, depending on the NLG requirements. Consequently, we have adapted some of these subtasks for the precipitation NLG module: content determination, sentence aggregation, lexicalization and linguistic realization. Others such as document planning were not considered, since in our case the NLG complexity is aimed at a sentence level. This process is summarized in Figure \ref{precnlg}.

\begin{figure}[!b]
\centering
\includegraphics[width=0.7\columnwidth]{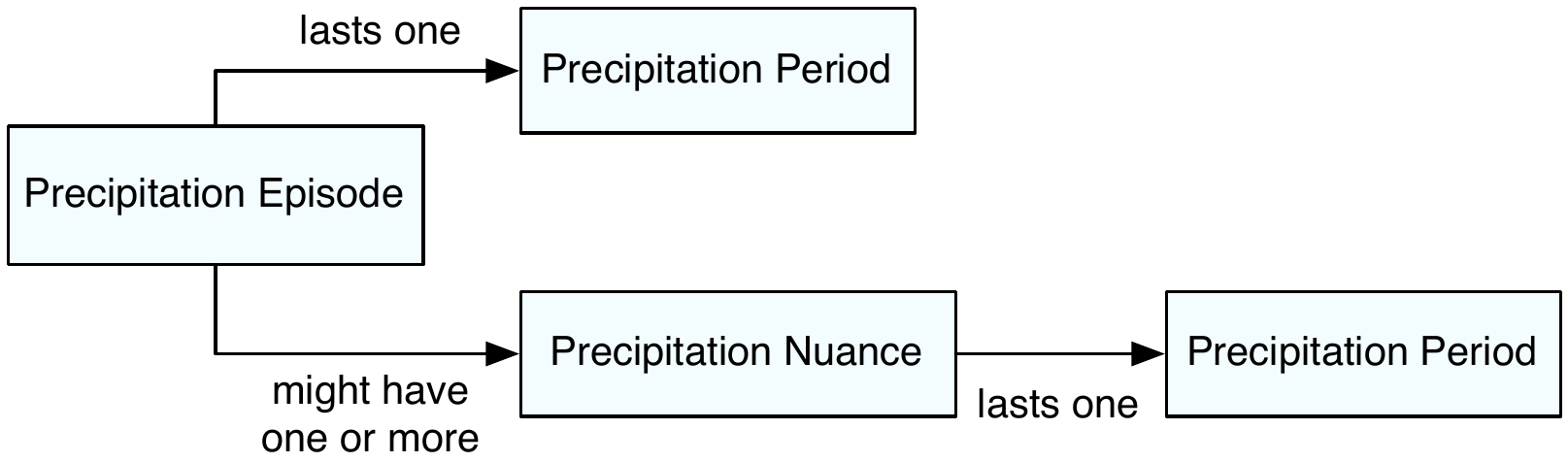}
\caption{Precipitation data object structure for the NLG stage.} 
\label{precstructure}
\end{figure}

Content determination is defined in \cite{bib_building_nlgart} as the process which decides what information should be communicated in the text. This is done by creating a set of data objects (messages) which contain the filtered and summarized data. In our method, this task is partially performed in the linguistic description stage by the precipitation operator, which extracts the relevant data from the raw data and converts it into an intermediate language. The remaining task is to convert the intermediate code into data objects, which is done by the precipitation NLG module parser. As a result, a list of precipitation episodes, whose structure is shown in Fig. \ref{precstructure}, is created and used by the subsequent natural language generation subtasks.

The precipitation data object structure in Fig. \ref{precstructure} shows that a precipitation episode has a duration (which can range from a single instant to the whole term). Furthermore, it might have associated nuances, which are subintervals within the episode in which the precipitation can be of different nature than rain (of snow, of hail or stormy).

The next NLG subtask we have adopted in our approach is sentence aggregation, which consists in grouping messages into sentences. We have contemplated three different ways of aggregating the precipitation episodes: by episodes, by days and whole-term aggregation. Consequently, we have created three different submodules which perform not only sentence aggregation, but also lexicalization and linguistic realization.

Lexicalization, which is the process of deciding which specific words and phrases should be chosen to express the concepts and relations in the messages, employs label sets defined in the NLG templates described in the previous approach.

Linguistic realization produces a text which is syntactically, morphologically and orthographically correct. Our precipitation approach obtains three candidate natural language precipitation sentences which describe the same input meteorological data set. The final output sentence for precipitation will be the shortest of the three, since we want to ensure that the obtained natural language forecasts remain as concise and brief as possible \cite{bib_felixisda}. 

\subsection{Implementation details}
This application has been developed in the cross-platform coding language Python, with the use of libraries for mathematical and fuzzy calculations (\textit{numpy}, \textit{pyfuzzy}) or text pattern recognition by grammars (\textit{pyparsing}). The current implementation supports both Linux and Windows systems.
The initially supported languages include Spanish and Galician. English was also included for research and scientific exposure purposes.

\section{Validation and results}
In this section we address the validation process for GALiWeather, which consists in an exhaustive expert-based revision and quality assessment of a set of automatically generated text forecasts obtained by the application. For this, we briefly discuss the state of the art in validation methodologies for both NLG and LDD fields and, based on these approaches, we explain in detail the validation methodology we have followed and its associated results. For illustration purposes, we present beforehand three examples of linguistic descriptions from the validation set obtained with the application.
\subsection{Examples of automatic weather forecasts}

\begin{figure}[!b]
\centering
\includegraphics[width=0.8\columnwidth]{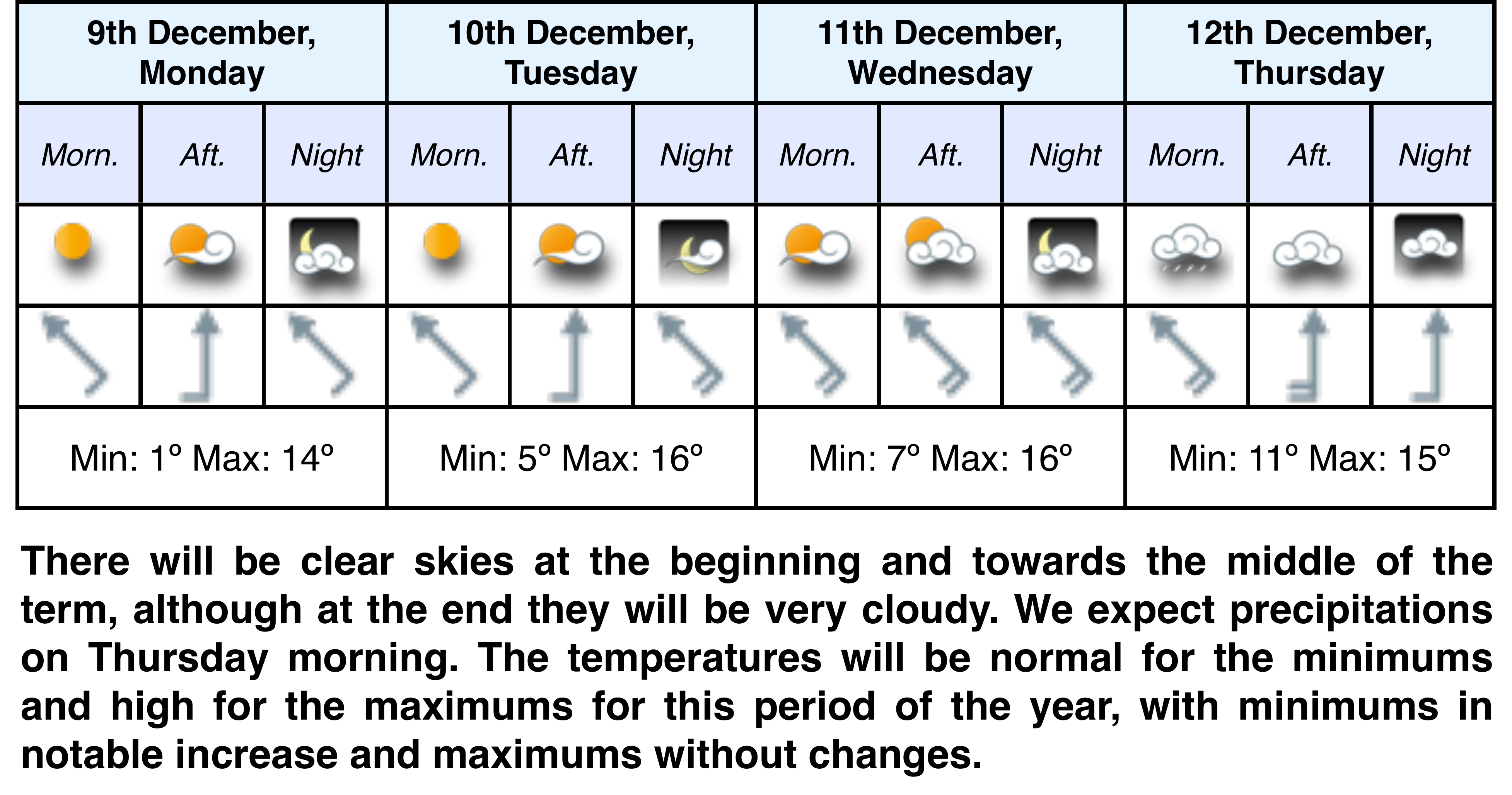}
\caption{Linguistic description forecast obtained with the application using real forecast data for Pontevedra, 9th of December, 2013.} 
\label{example3}
\end{figure}

Although the short-term prediction data series are limited to 32 values, the number of phenomena which must be considered and its temporal variability ensures a high richness in the obtained linguistic descriptions. As a proof of this richness, we present in this section the following examples covering several meteorological situations.

\begin{figure}[!t]
\centering
\includegraphics[width=0.8\columnwidth]{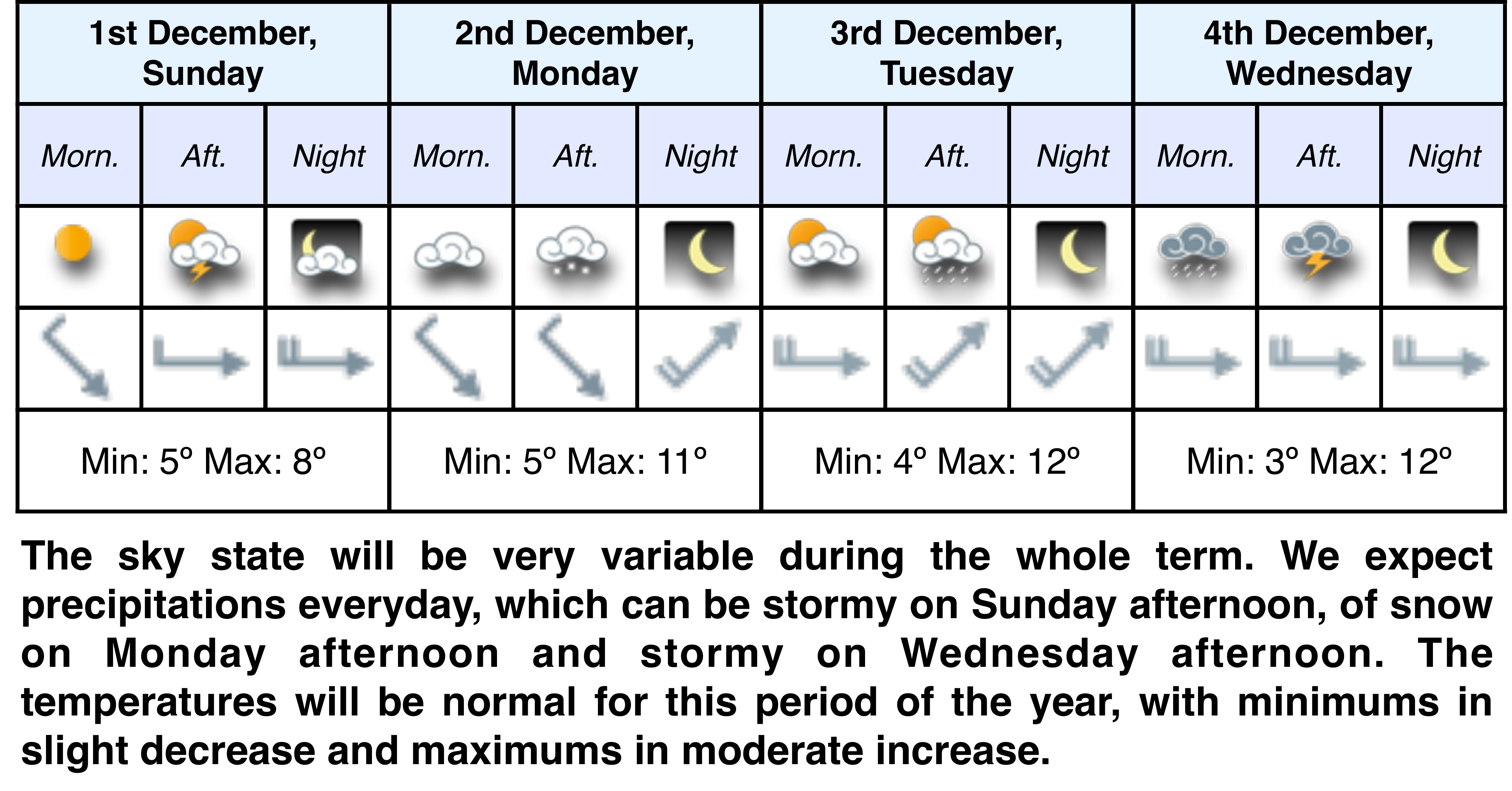}
\caption{Linguistic description forecast obtained with the application using synthetic data.} 
\label{example1}
\end{figure}

The example shown in Fig. \ref{example3} includes real forecast data for the town of Pontevedra, issued the 9th of December by MeteoGalicia. This case shows how GALiWeather performs in common meteorological situations, where the weather changes progressively.

\begin{figure}[!t]
\centering
\includegraphics[width=0.8\columnwidth]{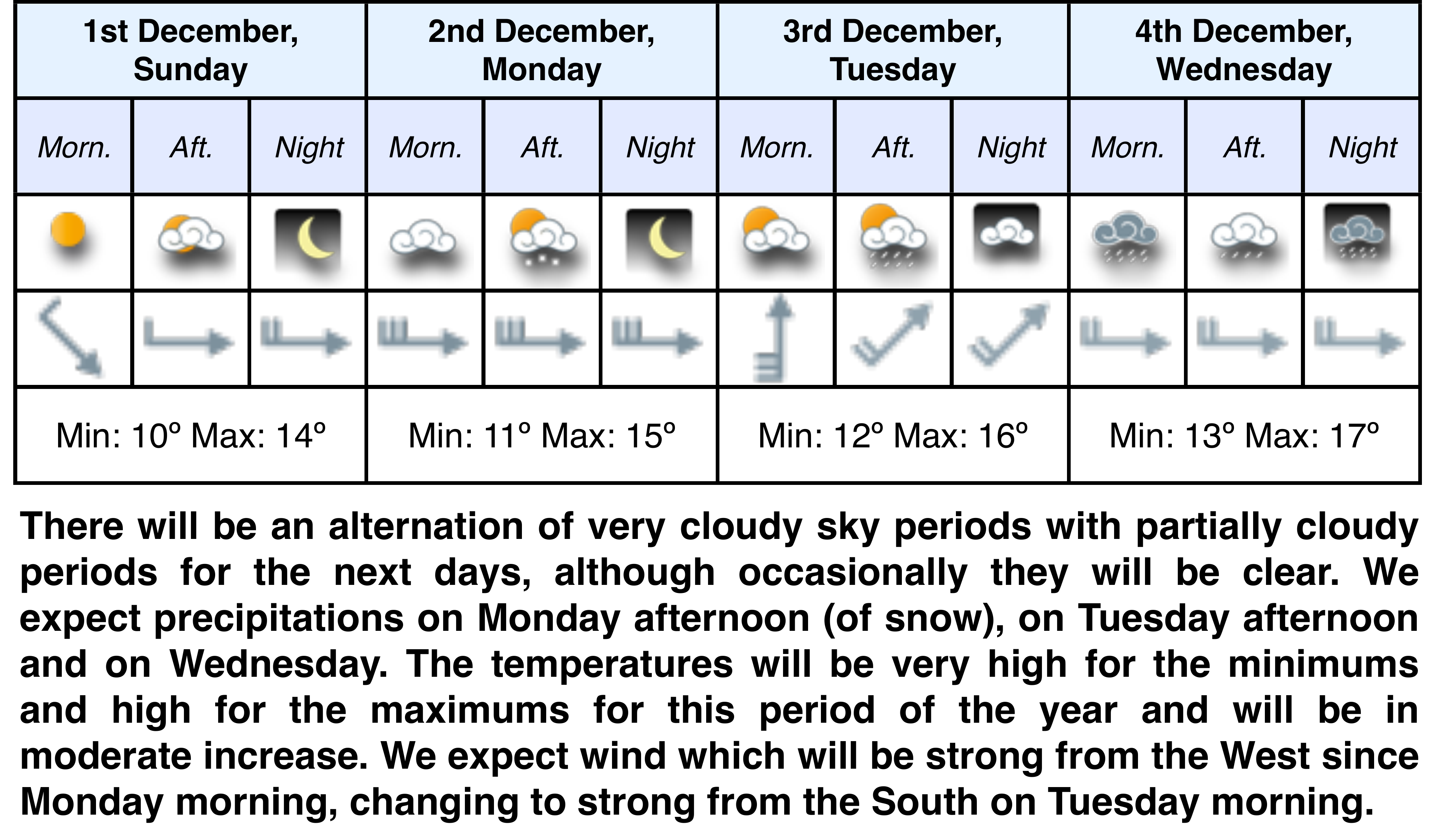}
\caption{Linguistic description forecast obtained with the application using synthetic data.} 
\label{example2}
\end{figure}

The examples shown in Fig. \ref{example1} and Fig. \ref{example2} present unusual and odd meteorological conditions, which were generated using synthetic data forecasts. These cases were created to test the application robustness under uncommon situations. Both examples include several meteorological phenomena, such as snow, storm, strong winds and temperature variations. Furthermore, each example shows a different precipitation sentence which aggregates the precipitation periods in a different way, as described in  Section III-C.

\subsection{Validation methodology}
Validating automatic natural language generated texts is still an open challenge, even within the NLG field \cite{bib_taskeval}. Several validation approaches do exist, both human and automatic, although in general, the human validation by experts is considered the most reliable \cite{bib_comparing}, \cite{bib_btnurse}. Consequently, the vast majority of NLG systems are validated using expert assessment, which usually implies answering questions about different aspects of the output texts. In the case of the LDD field several criteria have been proposed for evaluating and measuring the quality of the linguistic descriptions objectively \cite{bib_felixisda}, but they are not applicable in every approach and the information they provide is very limited compared to that of an expert, besides the fact that many LDD approaches do not reach the NLG stage and are not subject to a full validation process.

MeteoGalicia's meteorologists have provided support for a human expert validation of the results, which has allowed us to refine the proposed solution in a way that ensures it works under realistic conditions and cases. For this, we have performed the following validation process:
\begin{enumerate}
\item \textbf{Dataset collection creation}. A collection of 45 forecast datasets was created by the meteorologists. This collection includes synthetic and real forecast data, which covers common as well as unusual meteorologic scenarios, similar to the ones presented in the examples in Section IV-A.
\item \textbf{Natural language forecast automatic generation}. From this collection of forecast datasets, 45 automatically generated natural language forecasts were obtained.
\item \textbf{Polishing stage}. These 45 natural language forecasts generated by our application were evaluated by a meteorologist who assessed their quality taking into account their most relevant aspects and dimensions of interest. This initial evaluation was made to obtain preliminary conclusions and polish our approach in those aspects which needed to be improved.
\item \textbf{Natural language forecast automatic generation}. Once the changes to our approach were implemented, new 45 automatically generated language forecasts have been obtained from the original collection of forecast datasets. 
\item \textbf{Validation stage}. We have requested the expert to assess the new 45 automatically generated natural language forecasts. As opposed to the results from the polishing stage, which served to identify certain issues and potential improvements, the results of this stage allow to discern if the improvements in our approach are effective and, more importantly, if our application meets the expert's requirements and is consequently prepared to be released as a public service.
\end{enumerate}

\begin{figure}[!b]
\centering
\includegraphics[width=0.7\columnwidth]{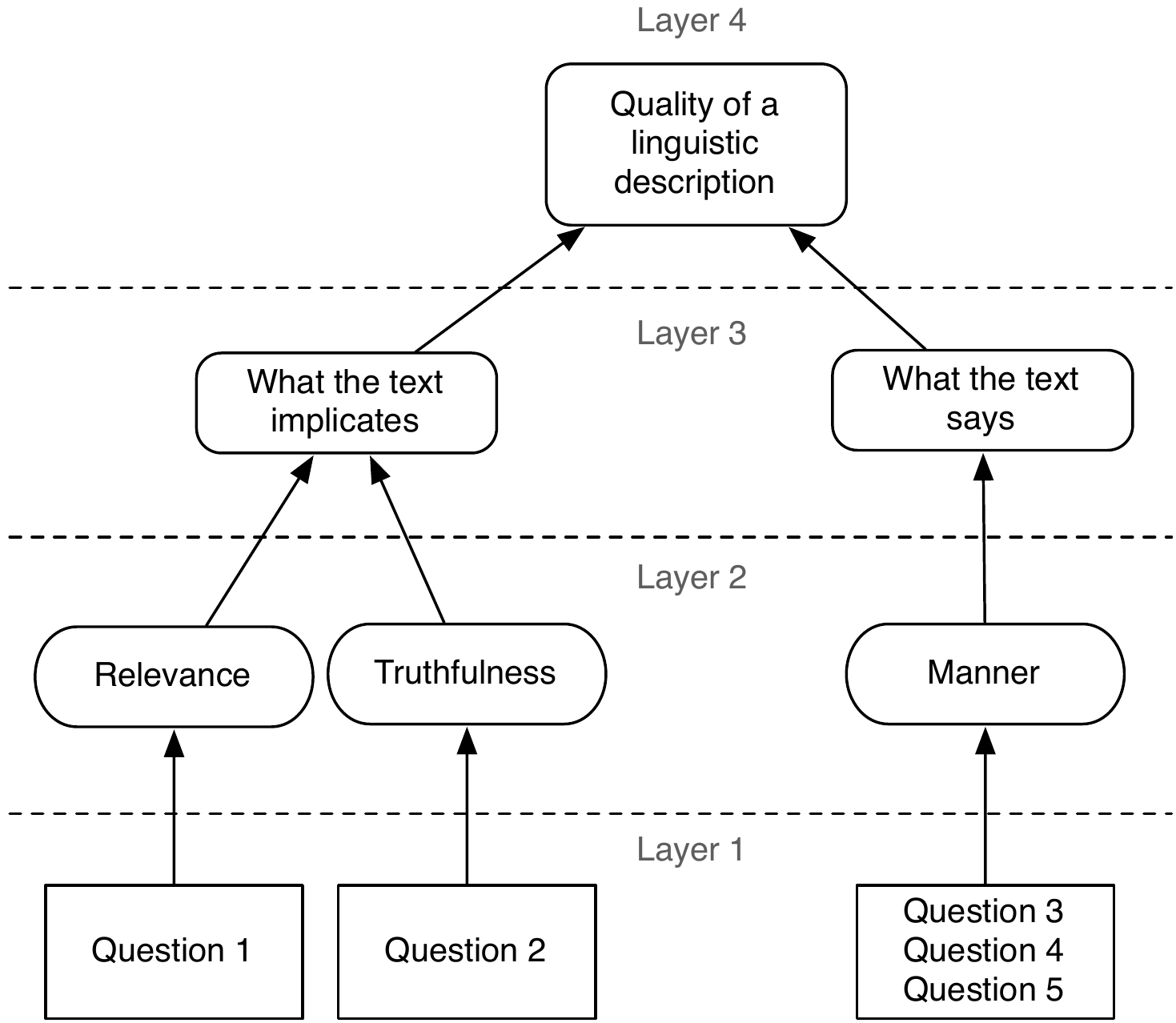}
\caption{Schema of the validation composition.} 
\label{evaldiagram}
\end{figure}

In order to assess the quality of the automatically generated forecasts, we have provided the expert meteorologist with a questionnaire which follows the approach presented in \cite{bib_quality}. This questionnaire covers three key dimensions about the generated weather forecasts, as shown in Fig. \ref{evaldiagram}:

\begin{itemize}
\item Relevance: Does the forecast include all the kind of information the expert would include?
\item Truthfulness: Does the included information in the forecast reflect the numeric-symbolic forecast correctly?
\item Manner: Does the forecast express the information properly? Is it well formatted?
\end{itemize}

These three dimensions are directly classified into two higher level categories, "what the text implicates" and "what the text says", which altogether determine the quality of the generated forecast. More specifically, the questionnaire we propose consists of five questions which deal in more depth with the previous three dimensions:
\begin{itemize}
\item \textbf{Question 1}: ``Indicate in which degree you identify the type of results expressed as the type of results expressed by yourself: a) For sky coverage b) For precipitations c) For wind d) For temperatures''.

This question determines the grade in which an expert identifies the generated forecast with the ones he creates. For reasons of precision, and in order to identify more specific issues in each forecast variable, Question 1 was divided into four subquestions, one for each forecast variable.
\item \textbf{Question 2}: ``Do you agree with the provided descriptions? a) For sky coverage b) For precipitations c) For wind d) For temperatures''.

This question considers the degree of truthfulness of the generated description, this is, the degree in which the content of the forecast reflects faithfully the information within the numeric-symbolic forecast data. Similar to Question 1, Question 2 is divided into four subquestions. With the ratings of Questions 1 and 2, we obtain the partial rating of the forecast related to ``what the text implicates''.
\item \textbf{Question 3}: ``Indicate in which degree the vocabulary is used correctly''.

This question evaluates if the vocabulary from the meteorology domain is used properly.
\item \textbf{Question 4}: ``Indicate in which degree the content is correctly grouped to facilitate the comprehension of the description''.

This question evaluates if the information in the natural language description is properly grouped and not repetitive.
\item \textbf{Question 5}: ``Indicate in which degree the format of the report, including the punctuation, is the most adequate''.

Question 5 considers aspects related to the forecast text presentation, such as punctuation. With the ratings of Questions 3, 4 and 5 we obtain the partial rating ``what the text says''.
\end{itemize}

Each of these questions must be answered as a number in a 1-5 scale (from 1 ``very negative'' to 5 ``very positive''). Thus, in order to calculate the global score for the collection of automatically generated forecasts, we follow the global aggregation schema defined in expression (2). Following this quality measure approach, the quality $Q$ of an automatically generated natural language weather forecast $S_i$ is defined as the arithmetic mean of the two dimensions in Layer 3 (Fig. \ref{evaldiagram}):

\begin{equation}
 Q_{S_i} = \frac{ \frac{\overline{p_1} + \overline{p_2}}{2}+\frac{p_3 +p_4+p_5}{3}}{2}
  \label{score_e}
\end{equation}

The terms $\overline{p_1}$ and $\overline{p_2}$ correspond to the average score of the subquestions a, b, c and d for Question 1 and Question 2, respectively. The remaining terms, $p_3$, $p_4$ and $p_5$ are the scores for Questions 3, 4 and 5. As \ref{score_e} shows, the average of $\overline{p_1}$ and $\overline{p_2}$ (``what the text implicates'') and the average of  $p_3$, $p_4$ and $p_5$ (``what the text says'') determine the quality of a forecast. Thus, the global quality score $GQ$ for our collection of automatically generated natural language forecasts is obtained as the average of the validation cases quality score: $GQ = \sum\limits_{i=1}^n \frac{Q_{S_i}}{n}$, where $n=45$ in our case.

\subsection{Results}
\begin{table}[!b]
\centering
\caption{Polishing stage questionnaire score}
\begin{tabular}{| l | c | c |}
\hline
Questions & Average score & Standard deviation  \\ \hline
\noalign{\smallskip} \hline
Q. 1 (a-d) & (3.6 3.93 5 4) & (0.45 0.75 0 0.57) \\ \hline
Q. 2 (a-d) & (4.04 4.44 5 4.86) & (0.36 0.5 0 0.34) \\ \hline
Q. 3 & 5 & 0 \\ \hline
Q. 4 & 3.64 & 0.77 \\ \hline
Q. 5 & 4.26 & 0.49 \\ \hline 
\textbf{$GQ$}  & \textbf{4.35} & \textbf{0.22} \\ \hline
\end{tabular}
\label{results}
\end{table}

One expert meteorologist answered the proposed questionnaire for the initial 45 automatically generated forecasts. Table \ref{results} shows that, in general, the meteorologist's assessment about the content of the forecasts was very positive for the initial test (with an average global score ($GQ$) of 4.35 out of 5 and a deviation of 0.22). In this sense, the expert identified the content and language of the generated forecasts with the ones he would provide in a high degree. However, from each individual question score we could extract additional conclusions which, in general, implied that there was room for improvement, especially on Question 4 and on some variables from Question 1 and 2. This was due to several repetitive sentences produced by the NLG stage in some of the variables (especially precipitation) and to some expressions which were not appropriate for some variables.

\begin{table}[!b]
\centering
\caption{Validation questionnaire score}
\begin{tabular}{| l | c | c |}
\hline
Questions & Average score & Standard deviation  \\ \hline
\noalign{\smallskip} \hline
Q. 1 (a-d) & (5 5 5 5) & (0 0 0 0) \\ \hline
Q. 2 (a-d) & (4.97 4.53 5 5) & (0.14 0.5 0 0) \\ \hline
Q. 3 & 5 & 0 \\ \hline
Q. 4 & 4.64 & 0.48 \\ \hline
Q. 5 & 4.53 & 0.50 \\ \hline 
\textbf{$GQ$} & \textbf{4.83} & \textbf{0.18} \\ \hline
\end{tabular}
\label{results2}
\end{table}

Based on the results obtained for the polishing stage, we have improved the NLG modules to address the issues found in our first approach and a validation test has been performed by the meteorologist with new 45 automatically generated natural language forecasts. With an average score of 4.83 out of 5 and a deviation of 0.18 (as Table \ref{results2} shows), the quality increase is substantial. In particular, the results in Question 1 show that the expert fully identifies the automatically generated forecasts as if they were produced manually by him. The fact that both content and language from the automatic forecasts are almost indistinguishable from those that an expert would produce are the most important among the several quality aspects which can be measured for a NLG approach. 
The remaining Questions also show increased scores compared to the first assessment.

\section{Application conceptualization}
\begin{figure}[!b]
\centering
\includegraphics[width=0.8\columnwidth]{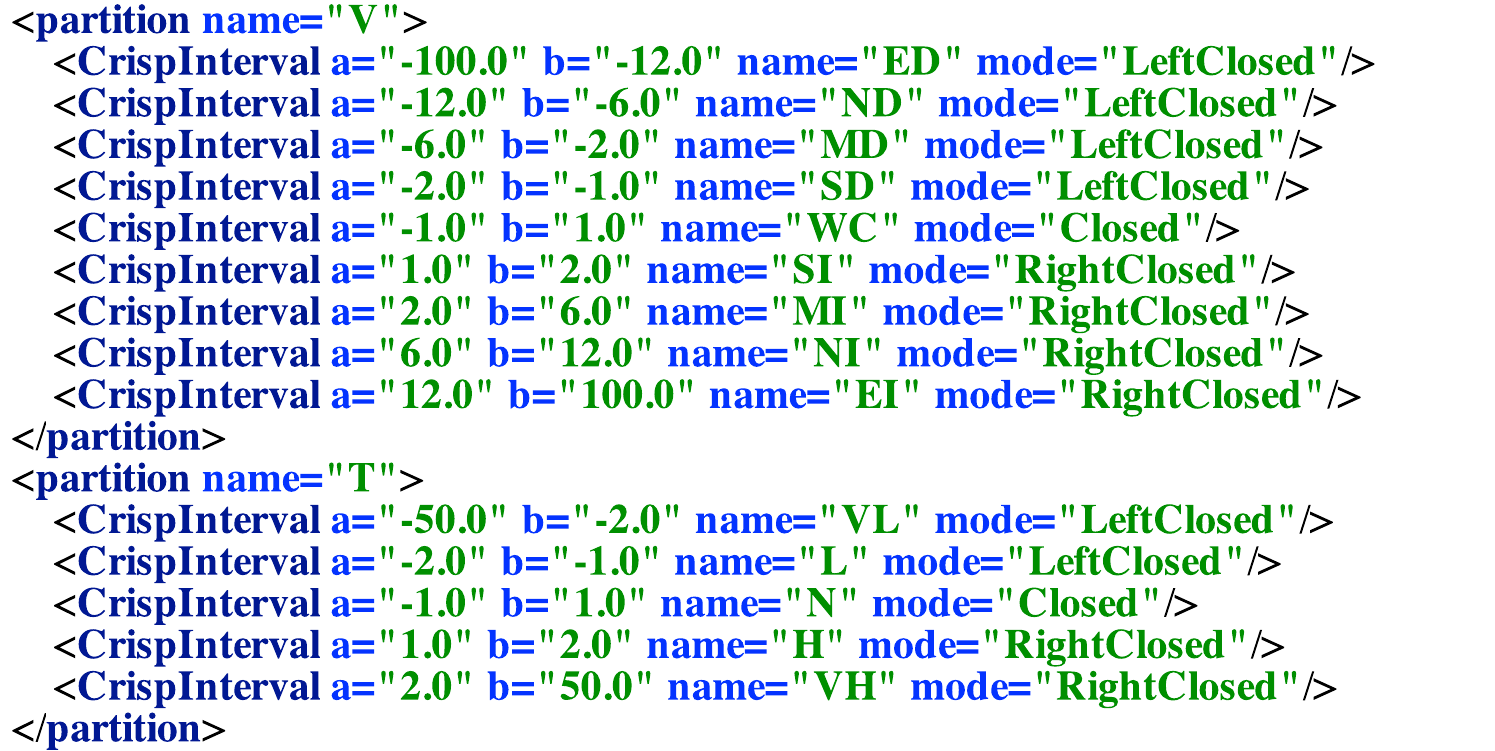}
\caption{Crisp partition sets for temperature variation and climatic behavior as defined in the configuration document.} 
\label{partitionsample}
\end{figure}

The solution we have presented addresses a specific practical problem by solving the need for providing 315 daily short term weather forecasts, which otherwise would not be possible to produce if they were manually created by a single meteorologist. As a consequence, the NLG stage is problem-oriented and is mostly not reusable. In spite of this, we want to stress the role that linguistic descriptions of data (LDD) techniques can play as a generic toolset which can be applied to many domains and give some insights into the generic methodology we are following for this LDD approach. For example, our application includes highly configurable linguistic description operators, which allow data series of any length and linguistic variables (implemented as fuzzy or crisp membership functions) with any number of labels as input. In fact, most of the changes made to improve the application during the whole development process were made to the linguistic variable definitions used by these operators (some of which are shown in Fig. \ref{partitionsample}) rather than to the operators themselves.

From our point of view, the main purpose of creating linguistic description solutions is to provide users with descriptions which make use of easily understandable familiar concepts found in natural language, imprecise and ambiguous in their nature. These concepts are usually modeled by employing some of the theoretical tools provided by the Computation With Perceptions field, such as fuzzy quantifiers, linguistic variables and others. However, the fact that these descriptions include linguistic terms neither implies they are actually expressed in natural language nor means they should be, as it occurs in NLG systems. In fact, both research fields seem rather complementary, in such a way that LDD provides tools for extracting the most relevant information in the form of (imprecise) linguistic terms, which then are used as an input to a NLG system to produce well-constructed sentences which are ready for human consumption. This is the approach we have followed in our solution, where LDD operators create input descriptions for an independent NLG system which generates natural language forecasts.

With a clearer view of which aspects LDD, in our opinion, should cover, we can abstract the basic elements which serve as pillars for a general LDD methodology. Many of the approaches described in the literature (e.g. those referred to in Section I) share several elements in common that can be taken into account for a flexible and reusable methodology for generating linguistic descriptions approaches:
\begin{itemize}
\item \textbf{Operators}. Operators extract information from raw data, converting numeric measurements into structures composed of linguistic terms. Originally, linguistic descriptions were conceived as quantified sentences, which resulted from applying fuzzy quantification models to data series. Therefore, many of the existing approaches use some kind of fuzzy quantification to obtain descriptions over one or several variables. For example, we can apply Zadeh's or other quantification models to produce a summary like ``Most days of the month were dry'' (in the case of rain data time series). However, many other operators which extract different pieces of information can be defined and implemented \cite{bib_felixisda}, such as: evaluation of a fuzzy label over the data series (e.g. ``Most of the temperatures were high in March''), search of data sequences fulfilling a given fuzzy label (e.g. ``Energy consumption was low between days 3 and 10''), search of increasing or decreasing patterns (e.g. ``There was a slight increase of valve pressure during the morning''), search of pitches in the dataset or of oscillation patterns (e.g. ``The system got unstable between 10:00 and 10:30''), event-counting operators (e.g. ``There were too many high pitches within the last hours'') or summarizing operators based on temporal/spatial hierarchies (e.g. ``The month was hot but the first week was cold''). For instance, for our LDD approach we have created highly configurable operators for each weather variable, according to the type of information that we needed to extract. These operators can be applied straight-forwardly to other variables by just replacing the partition sets for the current variables with partition sets for the new ones.
\item \textbf{Use of temporal/spatial hierarchies}. In the majority of cases, the numeric data series have an associated temporal and/or spatial component. This allows to arrange the data in hierarchies, which are usually defined by the experts in the application field. For example, in a temperature data series which covers one year, with one measurement per day, we can define a temporal hierarchy which would group the individual days in months, the months in seasons and so on. This considerably improves the exploitation of the available data, allowing to extract richer and more complex information. In our case, we have employed a time hierarchy which divides the short-term forecast temporal window into three subperiods for cloud coverage.
\item \textbf{Operator compositions}. Operators can be considered as the core primitives or atomic logical units of a framework which generates linguistic descriptions. These units can be combined in order to build more complex descriptions, depending on the requirements of the specific linguistic description problem. Therefore, means for mixing their outputs should be taken into account as additional elements in our framework. Our LDD approach does not make use of this concept, since the linguistic descriptions we obtain for each variable are independent.
\item \textbf{Evaluation criteria}. The raw output of linguistic description approaches usually consists of several candidate descriptions which must be filtered according to some pre-defined criteria, in order to ensure the quality and truthfulness of the selected final summary. Again, every specific problem needs its own set of adapted criteria, but also some general objective and reusable evaluation criteria, such as the description length, truth or fulfillment degree, data coverage, ambiguity, etc. should be used \cite{bib_felixisda}. In our case, we have employed the aggregation of fulfillment degrees of each fuzzy subperiod with respect to each cloud coverage label to obtain the best cloud coverage for each subperiod. Furthermore, we have also used the length of descriptions in order to discriminate the final precipitation text forecast.
\end{itemize}

Although all of these are concepts and notions taken from experience, we believe the main value of this methodology lies in the operators as the building blocks of the LDD approaches. If a collection of well-tested óboth in quality and usefulnessó operators for linguistic description of data is gathered, the viability of a generic LDD framework to create domain-specific approaches is highly ensured. In order to achieve this, we propose a feedback process which combines bottom-up and top-down approaches. On one hand, we believe that the best way to ensure the usefulness of the operators is to generalize specific solutions taken from real life problems and test them in other contexts. On the other hand, intuition-based operators can also be proposed and tested to check whether the information they produce is relevant to the experts. This loop which goes from concrete to abstract and then vice versa would help to improve in a correct direction the general LDD framework. 

\section{Conclusions and Future Work}
We have presented GALiWeather, an application which obtains textual short-term weather forecasts for the 315 municipalities in Galicia, using the real data provided by MeteoGalicia. As opposed to other linguistic descriptions approaches, this solution is based on an applied development in a realistic application, whose definition and structure is inspired by the linguistic descriptions research field by using both fuzzy and crisp operators which extract relevant information, and also by the natural language generation field.

Furthermore, the automatically generated textual forecasts were thoroughly evaluated by a meteorologist in order to assess the quality of their contents and to check whether his expert knowledge was included correctly. The obtained results show that the textual forecasts fulfill the expert's requirements in a very high degree (4.83 out of 5). GALiWeather is to be released as a real service in a very near future, since the application fully meets the meteorologists' requirements. The automatic linguistic descriptions will be displayed as a new information service at MeteoGalicia's website \cite{bib_meteogalicia}.

The main value of GALiWeather resides in its ability to cover and support a service of high interest for a wide number of users, which can only be provided by generating descriptions of data in an automatic manner, due to the high number of textual forecasts (315 in this case) which must be obtained.

In a longer term we are considering other application fields in which linguistic descriptions will prove useful. Among them, we have identified linguistic descriptions on information and decision support environmental systems as a promising research line, where not only linguistic descriptions for single location data are interesting, but also descriptions which geographically aggregate data in order to provide region-wide information. This is a complex challenge which will include the description of data in both time and space dimensions. This will lead us to develop a general model which can be applied to application fields in other areas.

\section*{Acknowledgments}
The authors would like to thank the editors and referees for their comments and suggestions, which have led to a substantial improvement in the paper quality. We would also like to thank CITIUS and MeteoGalicia for their support and for providing personal and material means for the development of this application.

\bibliographystyle{IEEEtran}

\vspace*{-2\baselineskip}
\begin{IEEEbiography}[{\includegraphics[width=1in,height=1.25in,clip,keepaspectratio]{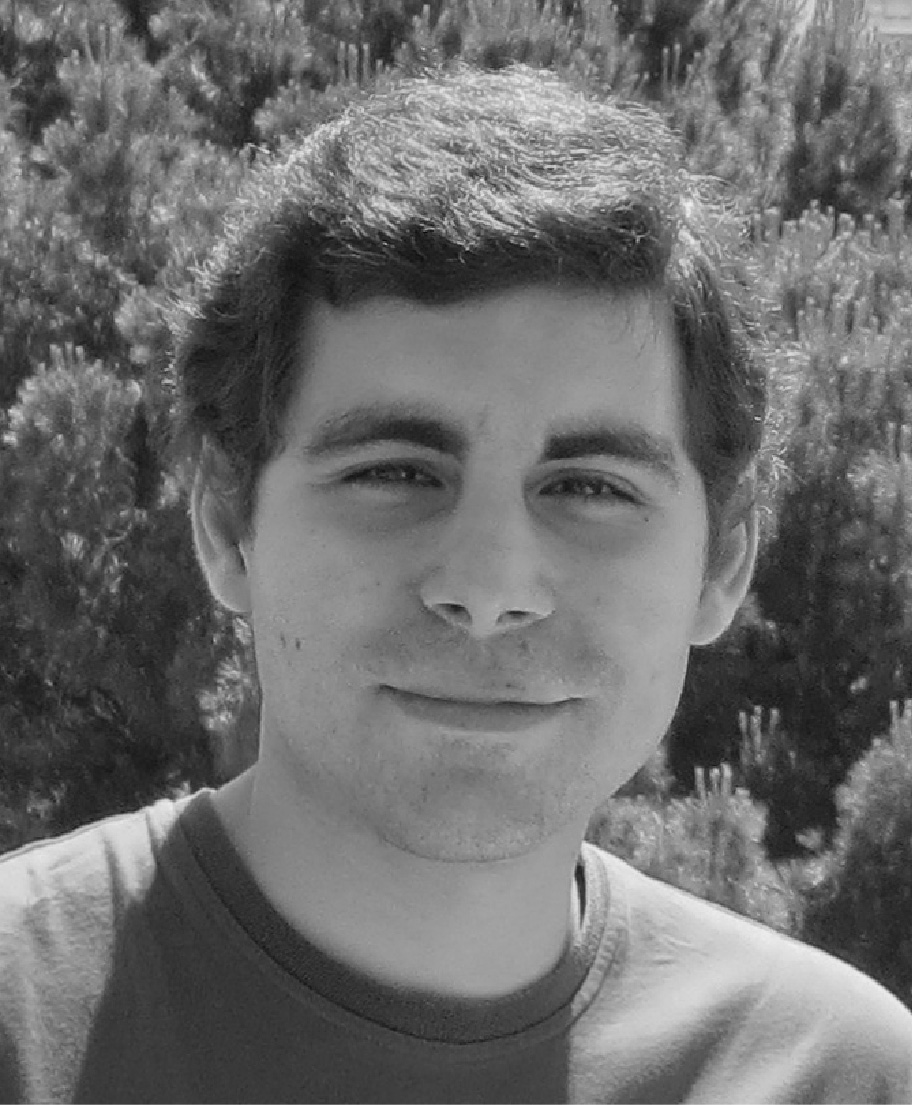}}]{Alejandro Ramos} received the M.S.c. degree in computer science from the University of Santiago de Compostela (USC), Spain, in 2011. He is currently a Ph.D. student at its Research Centre on Information Technologies (CiTIUS). His research interests include Linguistic Descriptions of Data and Natural Language Generation.
\end{IEEEbiography}
\vspace*{-5\baselineskip}
\begin{IEEEbiography}[{\includegraphics[width=1in,height=1.25in,clip,keepaspectratio]{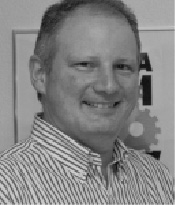}}]{Alberto J. Bugar\'in} received the Ph.D. degree in physics from the University of Santiago de Compostela (USC), Spain, in 1994. He is currently a Full Professor at its Research Centre on Information Technologies (CiTIUS). His research interests mainly focus on Linguistic Data Description of Data using Natural Language Generation, Machine Learning techniques for fuzzy knowledge bases discovery and Fuzzy Temporal knowledge representation and reasoning. On these and related topics and their applications he has published more than 150 scientific refereed papers and participated in more than 40 R+D projects and contracts.
\end{IEEEbiography}
\vspace*{-2\baselineskip}
\begin{IEEEbiography}[{\includegraphics[width=1in,height=1.25in,clip,keepaspectratio]{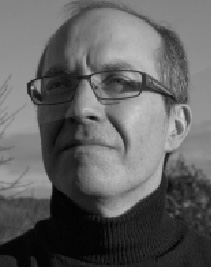}}]{Sen\'en Barro} received the Ph.D. in physics with distinction from the University of Santiago de Compostela (USC), Spain, in 1988.  He is Professor in the area of Computer Science and Artificial Intelligence. He was head of the Computer and Electronic Department of the University of Santiago de Compostela from 1993 to 2002, and the rector of this university from 2002 to 2010. Since May 2008 he is the president of RedEmprendia, which is made of 24 European and Latin American universities, focused on transfer on R\&D, innovation and entrepreneurship.
He founded the USC Intelligent Systems Group, which he also directs, and which currently has more than 40 members and is one of the first Artificial Intelligence groups in Spain.
He has been editor or author of seven books and author of more than 200 scientific articles. He has also been member of organizing, scientific and publishing committees of international conferences and journals.
\end{IEEEbiography}
\vspace*{-40\baselineskip}
\begin{IEEEbiography}[{\includegraphics[width=1in,height=1.25in,clip,keepaspectratio]{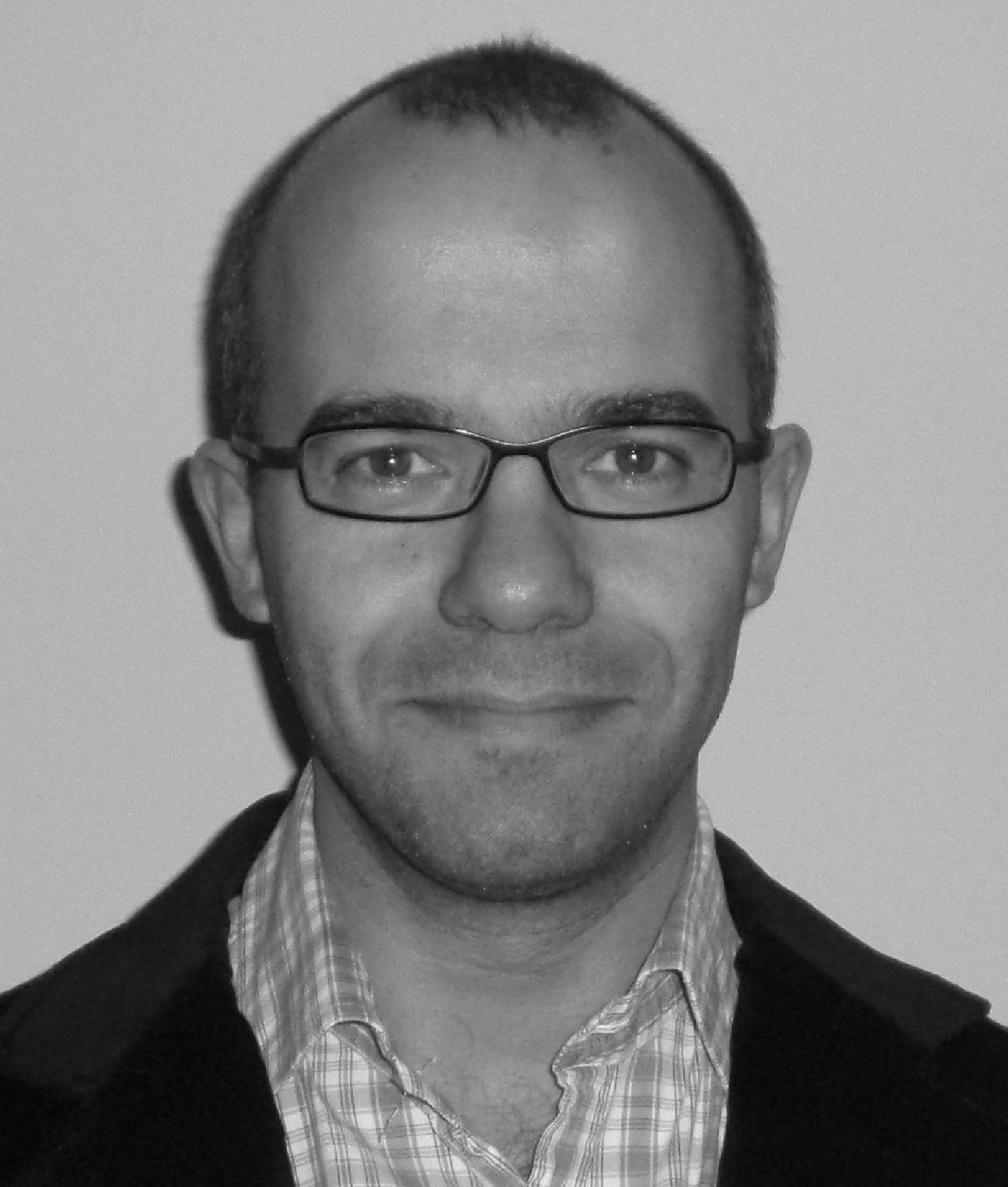}}]{Juan Taboada}
received the Ph.D. Degree in physics from the University of Santiago de Compostela (USC), Spain, in 1999, after a two-year stage in the University of Paris VI.
He currently leads the operational weather forecast department in MeteoGalicia, the Galician (NW Spain) Meteorological Agency. His research areas include climate variability and change, and seasonal and weather forecasting.
\end{IEEEbiography}

\end{document}